\documentclass{article}

\usepackage{microtype}
\usepackage{graphicx}
\usepackage{booktabs} % for professional tables
\usepackage{multirow}

\usepackage{hyperref}

\usepackage[accepted]{icml2024}

\usepackage{amsmath}
\usepackage{amssymb}
\usepackage{mathtools}
\usepackage{amsthm}

\usepackage{caption}
\usepackage{subcaption}

\usepackage[capitalize,noabbrev]{cleveref}

\usepackage{enumitem}
\theoremstyle{plain}
\newtheorem{theorem}{Theorem}[section]
\newtheorem{proposition}[theorem]{Proposition}

\theoremstyle{definition}
\newtheorem{definition}[theorem]{Definition}

\theoremstyle{remark}

\usepackage[textsize=tiny]{todonotes}

\newcommand{\blue}[1]{{\textcolor{blue}{#1}}}

\begin{document}

\twocolumn[
\icmltitle{Feasibility Consistent Representation Learning for Safe Reinforcement Learning}



\icmlsetsymbol{equal}{*}

\begin{icmlauthorlist}
\icmlauthor{Zhepeng Cen}{cmu}
\icmlauthor{Yihang Yao}{cmu}
\icmlauthor{Zuxin Liu}{cmu}
\icmlauthor{Ding Zhao}{cmu}
\end{icmlauthorlist}

\icmlaffiliation{cmu}{Carnegie Mellon University}

\icmlcorrespondingauthor{Zhepeng Cen}{zcen@andrew.cmu.edu}

\icmlkeywords{Safe reinforcement learning, Representation learning, ICML}

\vskip 0.3in
]



\printAffiliationsAndNotice{}  % leave blank if no need to mention equal contribution

\begin{abstract}
In the field of safe reinforcement learning (RL), finding a balance between satisfying safety constraints and optimizing reward performance presents a significant challenge. A key obstacle in this endeavor is the estimation of safety constraints, which is typically more difficult than estimating a reward metric due to the sparse nature of the constraint signals. To address this issue, we introduce a novel framework named Feasibility Consistent Safe Reinforcement Learning (FCSRL). This framework combines representation learning with feasibility-oriented objectives to identify and extract safety-related information from the raw state for safe RL. Leveraging self-supervised learning techniques and a more learnable safety metric, our approach enhances the policy learning and constraint estimation. Empirical evaluations across a range of vector-state and image-based tasks demonstrate that our method is capable of learning a better safety-aware embedding and achieving superior performance than previous representation learning baselines. The project website is available at \url{https://sites.google.com/view/FCSRL}.
\end{abstract}

%%%%% NEW MATH DEFINITIONS %%%%%

% \usepackage{amsmath,amsfonts,bm}

% Mark sections of captions for referring to divisions of figures
\newcommand{\figleft}{{\em (Left)}}
\newcommand{\figcenter}{{\em (Center)}}
\newcommand{\figright}{{\em (Right)}}
\newcommand{\figtop}{{\em (Top)}}
\newcommand{\figbottom}{{\em (Bottom)}}
\newcommand{\captiona}{{\em (a)}}
\newcommand{\captionb}{{\em (b)}}
\newcommand{\captionc}{{\em (c)}}
\newcommand{\captiond}{{\em (d)}}

% Highlight a newly defined term
\newcommand{\newterm}[1]{{\bf #1}}

% Figure reference, lower-case.
\def\figref#1{figure~\ref{#1}}
% Figure reference, capital. For start of sentence
\def\Figref#1{Figure~\ref{#1}}
\def\twofigref#1#2{figures \ref{#1} and \ref{#2}}
\def\quadfigref#1#2#3#4{figures \ref{#1}, \ref{#2}, \ref{#3} and \ref{#4}}
% Section reference, lower-case.
\def\secref#1{section~\ref{#1}}
% Section reference, capital.
\def\Secref#1{Section~\ref{#1}}
% Reference to two sections.
\def\twosecrefs#1#2{sections \ref{#1} and \ref{#2}}
% Reference to three sections.
\def\secrefs#1#2#3{sections \ref{#1}, \ref{#2} and \ref{#3}}
% Reference to an equation, lower-case.
\def\eqref#1{equation~\ref{#1}}
% Reference to an equation, upper case
\def\Eqref#1{Equation~\ref{#1}}
% A raw reference to an equation---avoid using if possible
\def\plaineqref#1{\ref{#1}}
% Reference to a chapter, lower-case.
\def\chapref#1{chapter~\ref{#1}}
% Reference to an equation, upper case.
\def\Chapref#1{Chapter~\ref{#1}}
% Reference to a range of chapters
\def\rangechapref#1#2{chapters\ref{#1}--\ref{#2}}
% Reference to an algorithm, lower-case.
\def\algref#1{algorithm~\ref{#1}}
% Reference to an algorithm, upper case.
\def\Algref#1{Algorithm~\ref{#1}}
\def\twoalgref#1#2{algorithms \ref{#1} and \ref{#2}}
\def\Twoalgref#1#2{Algorithms \ref{#1} and \ref{#2}}
% Reference to a part, lower case
\def\partref#1{part~\ref{#1}}
% Reference to a part, upper case
\def\Partref#1{Part~\ref{#1}}
\def\twopartref#1#2{parts \ref{#1} and \ref{#2}}

\def\ceil#1{\lceil #1 \rceil}
\def\floor#1{\lfloor #1 \rfloor}
\def\1{\bm{1}}
\newcommand{\train}{\mathcal{D}}
\newcommand{\valid}{\mathcal{D_{\mathrm{valid}}}}
\newcommand{\test}{\mathcal{D_{\mathrm{test}}}}

\def\eps{{\epsilon}}

% Random variables
\def\reta{{\textnormal{$\eta$}}}
\def\ra{{\textnormal{a}}}
\def\rb{{\textnormal{b}}}
\def\rc{{\textnormal{c}}}
\def\rd{{\textnormal{d}}}
\def\re{{\textnormal{e}}}
\def\rf{{\textnormal{f}}}
\def\rg{{\textnormal{g}}}
\def\rh{{\textnormal{h}}}
\def\ri{{\textnormal{i}}}
\def\rj{{\textnormal{j}}}
\def\rk{{\textnormal{k}}}
\def\rl{{\textnormal{l}}}
% rm is already a command, just don't name any random variables m
\def\rn{{\textnormal{n}}}
\def\ro{{\textnormal{o}}}
\def\rp{{\textnormal{p}}}
\def\rq{{\textnormal{q}}}
\def\rr{{\textnormal{r}}}
\def\rs{{\textnormal{s}}}
\def\rt{{\textnormal{t}}}
\def\ru{{\textnormal{u}}}
\def\rv{{\textnormal{v}}}
\def\rw{{\textnormal{w}}}
\def\rx{{\textnormal{x}}}
\def\ry{{\textnormal{y}}}
\def\rz{{\textnormal{z}}}

% Random vectors
\def\rvepsilon{{\mathbf{\epsilon}}}
\def\rvtheta{{\mathbf{\theta}}}
\def\rva{{\mathbf{a}}}
\def\rvb{{\mathbf{b}}}
\def\rvc{{\mathbf{c}}}
\def\rvd{{\mathbf{d}}}
\def\rve{{\mathbf{e}}}
\def\rvf{{\mathbf{f}}}
\def\rvg{{\mathbf{g}}}
\def\rvh{{\mathbf{h}}}
\def\rvu{{\mathbf{i}}}
\def\rvj{{\mathbf{j}}}
\def\rvk{{\mathbf{k}}}
\def\rvl{{\mathbf{l}}}
\def\rvm{{\mathbf{m}}}
\def\rvn{{\mathbf{n}}}
\def\rvo{{\mathbf{o}}}
\def\rvp{{\mathbf{p}}}
\def\rvq{{\mathbf{q}}}
\def\rvr{{\mathbf{r}}}
\def\rvs{{\mathbf{s}}}
\def\rvt{{\mathbf{t}}}
\def\rvu{{\mathbf{u}}}
\def\rvv{{\mathbf{v}}}
\def\rvw{{\mathbf{w}}}
\def\rvx{{\mathbf{x}}}
\def\rvy{{\mathbf{y}}}
\def\rvz{{\mathbf{z}}}

% Elements of random vectors
\def\erva{{\textnormal{a}}}
\def\ervb{{\textnormal{b}}}
\def\ervc{{\textnormal{c}}}
\def\ervd{{\textnormal{d}}}
\def\erve{{\textnormal{e}}}
\def\ervf{{\textnormal{f}}}
\def\ervg{{\textnormal{g}}}
\def\ervh{{\textnormal{h}}}
\def\ervi{{\textnormal{i}}}
\def\ervj{{\textnormal{j}}}
\def\ervk{{\textnormal{k}}}
\def\ervl{{\textnormal{l}}}
\def\ervm{{\textnormal{m}}}
\def\ervn{{\textnormal{n}}}
\def\ervo{{\textnormal{o}}}
\def\ervp{{\textnormal{p}}}
\def\ervq{{\textnormal{q}}}
\def\ervr{{\textnormal{r}}}
\def\ervs{{\textnormal{s}}}
\def\ervt{{\textnormal{t}}}
\def\ervu{{\textnormal{u}}}
\def\ervv{{\textnormal{v}}}
\def\ervw{{\textnormal{w}}}
\def\ervx{{\textnormal{x}}}
\def\ervy{{\textnormal{y}}}
\def\ervz{{\textnormal{z}}}

% Random matrices
\def\rmA{{\mathbf{A}}}
\def\rmB{{\mathbf{B}}}
\def\rmC{{\mathbf{C}}}
\def\rmD{{\mathbf{D}}}
\def\rmE{{\mathbf{E}}}
\def\rmF{{\mathbf{F}}}
\def\rmG{{\mathbf{G}}}
\def\rmH{{\mathbf{H}}}
\def\rmI{{\mathbf{I}}}
\def\rmJ{{\mathbf{J}}}
\def\rmK{{\mathbf{K}}}
\def\rmL{{\mathbf{L}}}
\def\rmM{{\mathbf{M}}}
\def\rmN{{\mathbf{N}}}
\def\rmO{{\mathbf{O}}}
\def\rmP{{\mathbf{P}}}
\def\rmQ{{\mathbf{Q}}}
\def\rmR{{\mathbf{R}}}
\def\rmS{{\mathbf{S}}}
\def\rmT{{\mathbf{T}}}
\def\rmU{{\mathbf{U}}}
\def\rmV{{\mathbf{V}}}
\def\rmW{{\mathbf{W}}}
\def\rmX{{\mathbf{X}}}
\def\rmY{{\mathbf{Y}}}
\def\rmZ{{\mathbf{Z}}}

% Elements of random matrices
\def\ermA{{\textnormal{A}}}
\def\ermB{{\textnormal{B}}}
\def\ermC{{\textnormal{C}}}
\def\ermD{{\textnormal{D}}}
\def\ermE{{\textnormal{E}}}
\def\ermF{{\textnormal{F}}}
\def\ermG{{\textnormal{G}}}
\def\ermH{{\textnormal{H}}}
\def\ermI{{\textnormal{I}}}
\def\ermJ{{\textnormal{J}}}
\def\ermK{{\textnormal{K}}}
\def\ermL{{\textnormal{L}}}
\def\ermM{{\textnormal{M}}}
\def\ermN{{\textnormal{N}}}
\def\ermO{{\textnormal{O}}}
\def\ermP{{\textnormal{P}}}
\def\ermQ{{\textnormal{Q}}}
\def\ermR{{\textnormal{R}}}
\def\ermS{{\textnormal{S}}}
\def\ermT{{\textnormal{T}}}
\def\ermU{{\textnormal{U}}}
\def\ermV{{\textnormal{V}}}
\def\ermW{{\textnormal{W}}}
\def\ermX{{\textnormal{X}}}
\def\ermY{{\textnormal{Y}}}
\def\ermZ{{\textnormal{Z}}}

% Vectors
\def\vzero{{\bm{0}}}
\def\vone{{\bm{1}}}
\def\vmu{{\bm{\mu}}}
\def\vtheta{{\bm{\theta}}}
\def\va{{\bm{a}}}
\def\vb{{\bm{b}}}
\def\vc{{\bm{c}}}
\def\vd{{\bm{d}}}
\def\ve{{\bm{e}}}
\def\vf{{\bm{f}}}
\def\vg{{\bm{g}}}
\def\vh{{\bm{h}}}
\def\vi{{\bm{i}}}
\def\vj{{\bm{j}}}
\def\vk{{\bm{k}}}
\def\vl{{\bm{l}}}
\def\vm{{\bm{m}}}
\def\vn{{\bm{n}}}
\def\vo{{\bm{o}}}
\def\vp{{\bm{p}}}
\def\vq{{\bm{q}}}
\def\vr{{\bm{r}}}
\def\vs{{\bm{s}}}
\def\vt{{\bm{t}}}
\def\vu{{\bm{u}}}
\def\vv{{\bm{v}}}
\def\vw{{\bm{w}}}
\def\vx{{\bm{x}}}
\def\vy{{\bm{y}}}
\def\vz{{\bm{z}}}

% Elements of vectors
\def\evalpha{{\alpha}}
\def\evbeta{{\beta}}
\def\evepsilon{{\epsilon}}
\def\evlambda{{\lambda}}
\def\evomega{{\omega}}
\def\evmu{{\mu}}
\def\evpsi{{\psi}}
\def\evsigma{{\sigma}}
\def\evtheta{{\theta}}
\def\eva{{a}}
\def\evb{{b}}
\def\evc{{c}}
\def\evd{{d}}
\def\eve{{e}}
\def\evf{{f}}
\def\evg{{g}}
\def\evh{{h}}
\def\evi{{i}}
\def\evj{{j}}
\def\evk{{k}}
\def\evl{{l}}
\def\evm{{m}}
\def\evn{{n}}
\def\evo{{o}}
\def\evp{{p}}
\def\evq{{q}}
\def\evr{{r}}
\def\evs{{s}}
\def\evt{{t}}
\def\evu{{u}}
\def\evv{{v}}
\def\evw{{w}}
\def\evx{{x}}
\def\evy{{y}}
\def\evz{{z}}

% Matrix
\def\mA{{\bm{A}}}
\def\mB{{\bm{B}}}
\def\mC{{\bm{C}}}
\def\mD{{\bm{D}}}
\def\mE{{\bm{E}}}
\def\mF{{\bm{F}}}
\def\mG{{\bm{G}}}
\def\mH{{\bm{H}}}
\def\mI{{\bm{I}}}
\def\mJ{{\bm{J}}}
\def\mK{{\bm{K}}}
\def\mL{{\bm{L}}}
\def\mM{{\bm{M}}}
\def\mN{{\bm{N}}}
\def\mO{{\bm{O}}}
\def\mP{{\bm{P}}}
\def\mQ{{\bm{Q}}}
\def\mR{{\bm{R}}}
\def\mS{{\bm{S}}}
\def\mT{{\bm{T}}}
\def\mU{{\bm{U}}}
\def\mV{{\bm{V}}}
\def\mW{{\bm{W}}}
\def\mX{{\bm{X}}}
\def\mY{{\bm{Y}}}
\def\mZ{{\bm{Z}}}
\def\mBeta{{\bm{\beta}}}
\def\mPhi{{\bm{\Phi}}}
\def\mLambda{{\bm{\Lambda}}}
\def\mSigma{{\bm{\Sigma}}}

% Tensor
% \DeclareMathAlphabet{\mathsfit}{\encodingdefault}{\sfdefault}{m}{sl}
% \SetMathAlphabet{\mathsfit}{bold}{\encodingdefault}{\sfdefault}{bx}{n}
\newcommand{\tens}[1]{\bm{\mathsfit{#1}}}
\def\tA{{\tens{A}}}
\def\tB{{\tens{B}}}
\def\tC{{\tens{C}}}
\def\tD{{\tens{D}}}
\def\tE{{\tens{E}}}
\def\tF{{\tens{F}}}
\def\tG{{\tens{G}}}
\def\tH{{\tens{H}}}
\def\tI{{\tens{I}}}
\def\tJ{{\tens{J}}}
\def\tK{{\tens{K}}}
\def\tL{{\tens{L}}}
\def\tM{{\tens{M}}}
\def\tN{{\tens{N}}}
\def\tO{{\tens{O}}}
\def\tP{{\tens{P}}}
\def\tQ{{\tens{Q}}}
\def\tR{{\tens{R}}}
\def\tS{{\tens{S}}}
\def\tT{{\tens{T}}}
\def\tU{{\tens{U}}}
\def\tV{{\tens{V}}}
\def\tW{{\tens{W}}}
\def\tX{{\tens{X}}}
\def\tY{{\tens{Y}}}
\def\tZ{{\tens{Z}}}

% Graph
\def\gA{{\mathcal{A}}}
\def\gB{{\mathcal{B}}}
\def\gC{{\mathcal{C}}}
\def\gD{{\mathcal{D}}}
\def\gE{{\mathcal{E}}}
\def\gF{{\mathcal{F}}}
\def\gG{{\mathcal{G}}}
\def\gH{{\mathcal{H}}}
\def\gI{{\mathcal{I}}}
\def\gJ{{\mathcal{J}}}
\def\gK{{\mathcal{K}}}
\def\gL{{\mathcal{L}}}
\def\gM{{\mathcal{M}}}
\def\gN{{\mathcal{N}}}
\def\gO{{\mathcal{O}}}
\def\gP{{\mathcal{P}}}
\def\gQ{{\mathcal{Q}}}
\def\gR{{\mathcal{R}}}
\def\gS{{\mathcal{S}}}
\def\gT{{\mathcal{T}}}
\def\gU{{\mathcal{U}}}
\def\gV{{\mathcal{V}}}
\def\gW{{\mathcal{W}}}
\def\gX{{\mathcal{X}}}
\def\gY{{\mathcal{Y}}}
\def\gZ{{\mathcal{Z}}}

% Sets
\def\sA{{\mathbb{A}}}
\def\sB{{\mathbb{B}}}
\def\sC{{\mathbb{C}}}
\def\sD{{\mathbb{D}}}
% Don't use a set called E, because this would be the same as our symbol
% for expectation.
\def\sF{{\mathbb{F}}}
\def\sG{{\mathbb{G}}}
\def\sH{{\mathbb{H}}}
\def\sI{{\mathbb{I}}}
\def\sJ{{\mathbb{J}}}
\def\sK{{\mathbb{K}}}
\def\sL{{\mathbb{L}}}
\def\sM{{\mathbb{M}}}
\def\sN{{\mathbb{N}}}
\def\sO{{\mathbb{O}}}
\def\sP{{\mathbb{P}}}
\def\sQ{{\mathbb{Q}}}
\def\sR{{\mathbb{R}}}
\def\sS{{\mathbb{S}}}
\def\sT{{\mathbb{T}}}
\def\sU{{\mathbb{U}}}
\def\sV{{\mathbb{V}}}
\def\sW{{\mathbb{W}}}
\def\sX{{\mathbb{X}}}
\def\sY{{\mathbb{Y}}}
\def\sZ{{\mathbb{Z}}}

% Entries of a matrix
\def\emLambda{{\Lambda}}
\def\emA{{A}}
\def\emB{{B}}
\def\emC{{C}}
\def\emD{{D}}
\def\emE{{E}}
\def\emF{{F}}
\def\emG{{G}}
\def\emH{{H}}
\def\emI{{I}}
\def\emJ{{J}}
\def\emK{{K}}
\def\emL{{L}}
\def\emM{{M}}
\def\emN{{N}}
\def\emO{{O}}
\def\emP{{P}}
\def\emQ{{Q}}
\def\emR{{R}}
\def\emS{{S}}
\def\emT{{T}}
\def\emU{{U}}
\def\emV{{V}}
\def\emW{{W}}
\def\emX{{X}}
\def\emY{{Y}}
\def\emZ{{Z}}
\def\emSigma{{\Sigma}}

% entries of a tensor
% Same font as tensor, without \bm wrapper
\newcommand{\etens}[1]{\mathsfit{#1}}
\def\etLambda{{\etens{\Lambda}}}
\def\etA{{\etens{A}}}
\def\etB{{\etens{B}}}
\def\etC{{\etens{C}}}
\def\etD{{\etens{D}}}
\def\etE{{\etens{E}}}
\def\etF{{\etens{F}}}
\def\etG{{\etens{G}}}
\def\etH{{\etens{H}}}
\def\etI{{\etens{I}}}
\def\etJ{{\etens{J}}}
\def\etK{{\etens{K}}}
\def\etL{{\etens{L}}}
\def\etM{{\etens{M}}}
\def\etN{{\etens{N}}}
\def\etO{{\etens{O}}}
\def\etP{{\etens{P}}}
\def\etQ{{\etens{Q}}}
\def\etR{{\etens{R}}}
\def\etS{{\etens{S}}}
\def\etT{{\etens{T}}}
\def\etU{{\etens{U}}}
\def\etV{{\etens{V}}}
\def\etW{{\etens{W}}}
\def\etX{{\etens{X}}}
\def\etY{{\etens{Y}}}
\def\etZ{{\etens{Z}}}

% The true underlying data generating distribution
\newcommand{\pdata}{p_{\rm{data}}}
% The empirical distribution defined by the training set
\newcommand{\ptrain}{\hat{p}_{\rm{data}}}
\newcommand{\Ptrain}{\hat{P}_{\rm{data}}}
% The model distribution
\newcommand{\pmodel}{p_{\rm{model}}}
\newcommand{\Pmodel}{P_{\rm{model}}}
\newcommand{\ptildemodel}{\tilde{p}_{\rm{model}}}
% Stochastic autoencoder distributions
\newcommand{\pencode}{p_{\rm{encoder}}}
\newcommand{\pdecode}{p_{\rm{decoder}}}
\newcommand{\precons}{p_{\rm{reconstruct}}}

\newcommand{\laplace}{\mathrm{Laplace}} % Laplace distribution

\newcommand{\E}{\mathbb{E}}
\newcommand{\Ls}{\mathcal{L}}
\newcommand{\R}{\mathbb{R}}
\newcommand{\emp}{\tilde{p}}
\newcommand{\lr}{\alpha}
\newcommand{\reg}{\lambda}
\newcommand{\rect}{\mathrm{rectifier}}
\newcommand{\softmax}{\mathrm{softmax}}
\newcommand{\sigmoid}{\sigma}
\newcommand{\softplus}{\zeta}
\newcommand{\KL}{D_{\mathrm{KL}}}
\newcommand{\TV}{D_{\mathrm{TV}}}
\newcommand{\Var}{\mathrm{Var}}
\newcommand{\standarderror}{\mathrm{SE}}
\newcommand{\Cov}{\mathrm{Cov}}
% Wolfram Mathworld says $L^2$ is for function spaces and $\ell^2$ is for vectors
% But then they seem to use $L^2$ for vectors throughout the site, and so does
% wikipedia.
\newcommand{\normlzero}{L^0}
\newcommand{\normlone}{L^1}
\newcommand{\normltwo}{L^2}
\newcommand{\normlp}{L^p}
\newcommand{\normmax}{L^\infty}

\newcommand{\parents}{Pa} % See usage in notation.tex. Chosen to match Daphne's book.

\let\ab\allowbreak

\vspace{-3mm}
\section{Introduction}

Reinforcement Learning (RL) has achieved remarkable success in various domains, leveraging its capability to learn optimal policies by interacting with the environment. This success has spanned from mastering complex games \cite{mnih2015human, silver2016mastering, hessel2018rainbow} to enabling autonomous systems \cite{kiran2021deep, li2022metadrive, ding2023survey}. However, as RL applications venture into more critical areas such as healthcare, finance, and self-driving vehicles, ensuring safety alongside task performance becomes equally or even more imperative \cite{garcia2015comprehensive}.
Safe RL aims to learn a constraint satisfaction policy either by interacting with the environment or from static offline datasets to reduce the risk of policy deployment in safety-critical scenarios \cite{brunke2021safe}.

While many strategies have been explored for safe RL problem, from model-based approaches that predict and mitigate potential risks~\cite{kaiser2019model, as2022constrained}, to constrained optimization-based methods~\cite{zhang2020first, chow2018risk, yang2020projection} that ensure policy updates within the feasible set, there still remain several significant hurdles. Among them, one main challenge is cost estimation~\cite{achiam2017constrained, tessler2018reward}, which stems from two main sources: (1) the complexity of the raw state often complicates the prediction of single-step costs; and (2) the sparsity of cost signal intensifies the non-smoothness of ground-truth value function and causes larger noise in value estimation, which is similar to the cases in sparse-reward and goal-conditioned RL~\cite{andrychowicz2017hindsight, riedmiller2018learning, hare2019dealing, liu2022goal, eysenbach2022contrastive, cen2024learning}. The former influences the evaluation of immediate state safety while the second impacts the estimation of long-term safety. These issues cause a significant over-estimation or under-estimation of agent on costs. Therefore, the safe RL agent struggles to balance the objectives of reward maximization and constraint satisfaction, thus obtaining a suboptimal policy performance.

We propose Feasibility Consistent Safe Reinforcement Learning (FCSRL) to employ representation learning~\cite{bengio2013representation} to tackle above challenge. Inspired by the enhancement of dynamics-based representation learning on RL~\cite{yang2021representation, fujimoto2023sale}, we also leverage transition dynamics of environment to learn the underlying structure of the transition dynamics by applying self-supervised learning~\cite{chen2021exploring, he2020momentum} loss on adjacent states in the sample trajectory. Additionally, due to the sparsity of cost, the representation loss based on traditional metrics (e.g., value function) does not boost the safety-aware embedding learning, although they exhibit notable improvement in standard RL setting~\cite{ye2021mastering, hessel2021muesli, farquhar2021self}. To address this, we introduce a novel learning objective, feasibility score, with smoother nature compared to other cost metrics and adopt it as an auxiliary task in representation learning to refine the safety context features, providing a more accurate constraint estimation and a better trade-off between reward maximization and safety constraint satisfaction for policy learning.

The contributions of this paper can be summarized as: 
\begin{itemize}
    \item We demonstrate the smoothness of the adopted feasibility metric with both theoretical and empirical analysis. 
    \item We apply feasibility metric in representation learning to tackle the cost estimation challenge and propose a representation learning framework, which is compatible with most existing model-free safe RL algorithms.
    \item The extensive experiments on vector-state and image-based tasks demonstrate that the proposed method learns a better safety-aware embedding and exhibits remarkable and consistent advantages over previous representation learning methods, especially in the case of more stringent constraint. 
\end{itemize}

\section{Related Work}

\textbf{Safe reinforcement learning}. Safe RL aims to learn an optimal policy by maximizing the reward performance while satisfying the safety constraint~\cite{garcia2015comprehensive, achiam2017constrained, wachi2020safe, gu2022review, xu2022trustworthy, liu2022robustness} and another line of research is on safe exploration to improve the safety during training~\cite{sui2015safe, dalal2018safe, wachi2018safe, sootla2022saute, wang2023enforcing}. One common solution to the constrained optimization is primal-dual framework~\cite{ray2019benchmarking, ding2020natural} and solve an unconstrained optimization with a Lagrangian multiplier~\cite{chow2018risk}. Gradient-based update methods~\cite{tessler2018reward, zhang2020first} tunes the Lagrangian multiplier to maximize reward while satisfying constraint. Furthermore, \citet{stooke2020responsive} propose PID-based Lagrangian update to reduce the instability of cost; \citet{liu2022constrained} and \citet{huang2022constrained} apply variational inference to solve optimal multiplier directly, which exhibits better stability and performance during training~\cite{yao2023constraint}. Recent works incorporate model-based RL to improve data efficiency and final performance~\cite{berkenkamp2017safe, as2022constrained, huang2023safe}, but typically they require a much larger model to parameterize the environment dynamics.

% 1. primal-dual; pid lagrangian for stability; variational method, extended to distributional setting~\cite{huang2022constrained};
% 2. another line is model-based safe RL,

\textbf{Representation learning in RL}. In context of RL, representation learning typically refers to learning an abstraction or latent embedding or extracting a features of state or action space~\cite{lesort2018state, abel2018state, lesort2018state}, which is also applied in offline learning setting~\cite{yang2021representation, lin2024safety}. When state is high-dimensional (e.g., image-based input), it also involves compression from large raw state to a smaller latent vector~\cite{finn2016deep, liu2021return}. The advance in self-supervised learning~\cite{chen2020simple, he2020momentum, grill2020bootstrap, chen2021exploring} also inspires the representation learning to employ self-supervised loss on augmented states sourced from the same state~\cite{laskin2020curl, kostrikov2020image} or two temporally adjacent states to capture the embedding from dynamics structure ~\cite{schwarzer2020data, yang2021representation, fujimoto2023sale}. The idea of latent state representations via dynamics model is also closely related to the model-based RL, which also adopt latent embedding learning to improve the learning of world model~\cite{gelada2019deepmdp, hafner2019learning, hafner2020mastering, kaiser2019model, schrittwieser2020mastering, ye2021mastering}. The model-based RL requires a more expressive parameterization of dynamics model to provide an accurate imagination for planning. In addition to learning on dynamics or world model, another supervision signal for representation learning is value consistency~\cite{oh2017value, farahmand2017value}. \citet{grimm2020value} propose the value equivalence principle to enforce the value function prediction and Bellman backup by latent representation to align with the real environment model, which can improve the representation learning and value learning in recent works~\cite{schrittwieser2020mastering, hessel2021muesli, farquhar2021self, yue2023value}.
\section{Preliminaries}
\textbf{Safe reinforcement learning}. 
Safe RL can be formulated in the framwork of Constrained Markov Decision Process (CMDP), which is defined by the tuple $\gM = \langle \gS, \gA, \gT, r, c, \gamma, \mu_0 \rangle$~\cite{altman1999constrained}, where $\gS$ represents the state space, $\gA$ is the action space, $\gT:\gS \times \mathcal{A} \times \gS \xrightarrow{} [0, 1]$ is the transition function, $\mu_0: \gS \xrightarrow[]{} [0,1]$ is the initial state distribution, $\gamma$ is the discount factor, $r:\gS \times \gA \to \mathbb{R}$ is the reward function, and $c:\gS \times \gA \to \mathbb{R}$ is the cost to characterize the constraint. Typically, the cost signal serves as an indicator of the safety of the current state, which is much sparser than reward. Therefore, in this paper, we focus on the case of binary cost, i.e., $c(s,a)=1$ indicates the state-action pair $(s,a)$ is unsafe and $c(s,a)=0$ means it is safe. The objective of safe RL is to find the optimal policy within the constraints 
\begin{equation}
\max_{\pi}  J_r(\pi), \quad s.t. \quad J_c(\pi) \leq \epsilon
\end{equation}
where $J_{\boldsymbol{f}}(\pi) = \mathbb{E}_{\rho \sim \pi, s_0 \sim \mu_0}[ \sum_{t=0}^\infty \gamma^t {\boldsymbol{f}}(s,a) ], {\boldsymbol{f}}\in \{r, c\}$ is the reward or cost return of policy $\pi$; $\rho$ denotes the sampled trajectory; $\epsilon$ is the pre-defined constraint threshold. RL algorithms commonly adopt V value $V_{\boldsymbol{f}}^{\pi}(s)=\E_\pi[\sum_{t=0}^\infty \gamma^t {\boldsymbol{f}}(s,a)|s_0=s]$ and Q value function $Q_{\boldsymbol{f}}^{\pi}(s,a)=\E_\pi[\sum_{t=0}^\infty \gamma^t {\boldsymbol{f}}(s,a)|s_0=s,a_0=a]$ during training.

% $V_{\boldsymbol{f}}^\pi(\mu_0) = \mathbb{E}_{\tau \sim \pi, s_0 \sim \mu_0}[ \sum_{t=0}^\infty \gamma^t {\boldsymbol{f}}_t ], {\boldsymbol{f}}\in \{r, c\}$, is the value function that is the expectation of discounted return under the policy $\pi$ and the initial state distribution $\mu_0$.

To solve the constrained optimization problem, a common practice for safe RL~\cite{tessler2018reward, ray2019benchmarking, zhang2020first} is to transform it to an unconstrained one by introducing a Lagrangian multiplier $\lambda$:
\begin{equation}
    \min_{\lambda \geq 0} \max_\pi J_r(\pi) - \lambda (J_c(\pi) - \epsilon),
\end{equation}
which can be solved in a general primal-dual framework.

\textbf{Representation learning for safe RL}. 
In this paper, the main objective of representation learning is to learn a mapping from raw state $s$ to a latent embedding $z$ to facilitate RL part. Particularly, we aim to extract safety-related features in state space and distinctly manifest it within the embedding space. By utilizing this representation as input, we can learn a better policy with both task utility efficiency and constraint satisfaction performances.

% % In context of RL, the objective of representation learning is to discover an embedding of a state or state-action pair, which can capture the relava.
% encoder: $f:\gS\to \gZ$.
% the value equivalence principle  and value consistent models has proven to improve the representation learning and reward performances in both model-free and model-based RL settings. In context of model-free RL, the key idea of value consistency is to enforce the learned state representation $z$ to be able to predict the value function of the corresponding state $s$, i.e., $\tilde{v}(z) = V(s)$, where $z=h(s)$ and $\tilde{v}: \gZ\to\sR$ is the value prediction head in embedding learning. 
\section{Method}
% \textcolor{red}{The proposed method is compatible with general primal-dual algorithms(?)}

\begin{figure}[ht]
% \vspace{-5pt} 
\centering
\includegraphics[width=0.95\linewidth]{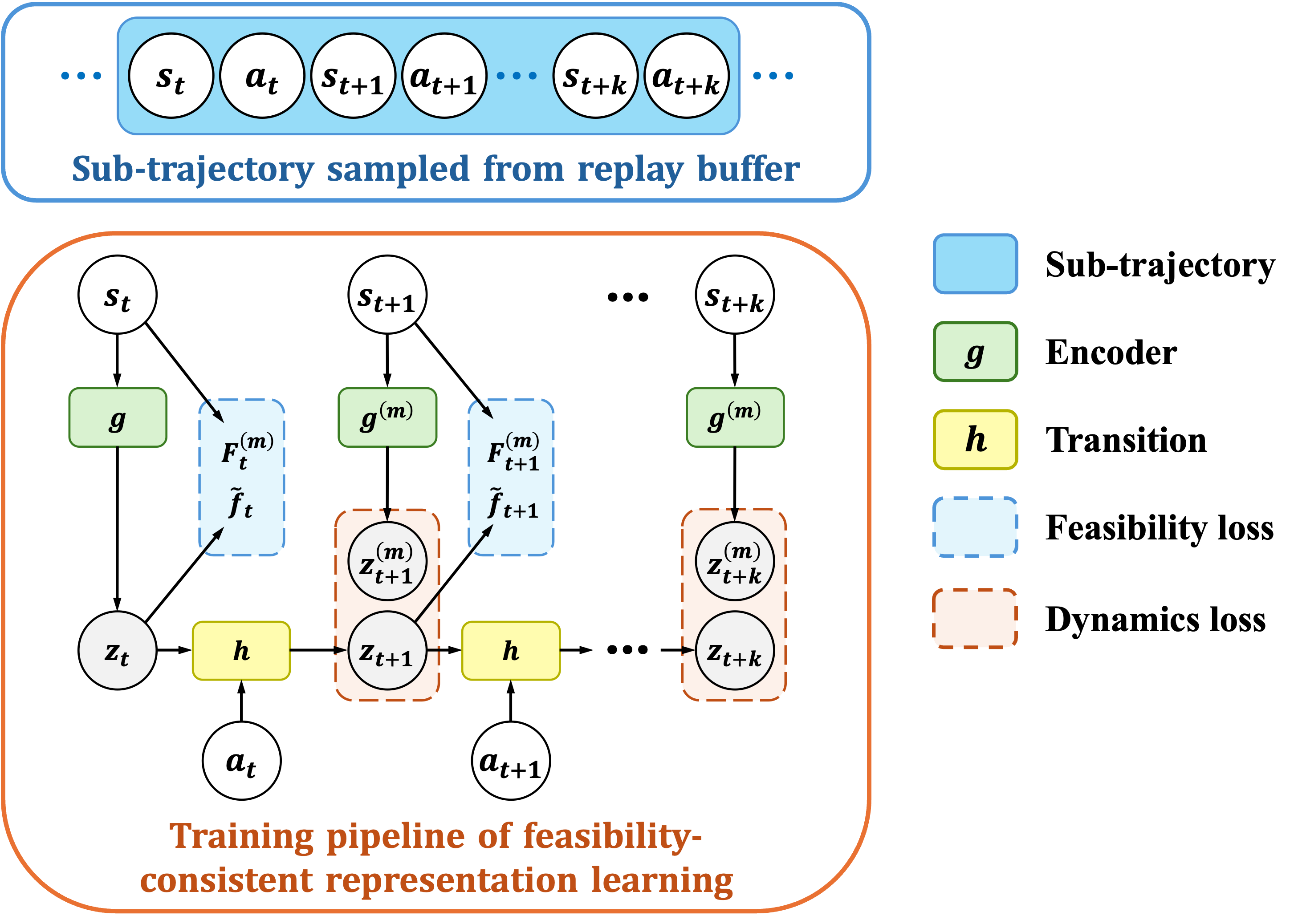}
\caption{The pipeline of feasibility consistent representation learning. There are two main parts in learning objective: (1) the dynamics loss is between the predicted representation $z_t$ and target $z_t^{(m)}$, and (2) the feasibility consistency loss is between $\tilde{f}_t$ predicted from representation $z$ and target feasibility $F^{(m)}$ estimated by Bellman bootstrap.
}
% \vspace{-10pt}
\label{fig:framework}
\end{figure}

% , which uses the feasbility of state as objective and generates a safety-aware embedding to facilitate safe reinforcement learning. 
In this section, we present \textbf{feasibility consistent} representation learning. Fig.~\ref{fig:framework} show the overview of representation learning pipeline, which includes following two main components.

\subsection{Transition Dynamics Consistency}
The representation learning achieves great success in non-sequential tasks and improves the sample efficiency of vision-input RL by mapping a high-dimensional observation to a low-dimensional embedding~\cite{laskin2020curl}. However, such mapping is not explicitly related to transition dynamics of the environment, which may be less helpful for decision-making. Therefore, based on the intuition that \textit{a good state embedding in RL should be predictive of the future}~\cite{schwarzer2020data, fujimoto2023sale}, we adopt dynamics consistency loss in representation learning. 

Given a transition pair $\tau = (s_t,a_t,s_{t+1})$ from replay buffer $\gD$, we use an encoder $g$ to encode the state $s_t$ to embedding $z_t=g(s_t)$ and a transition model $h$ to output $z_{t+1}=h(z_t,a_t)$ as the prediction of the next-state embedding $z_{t+1}$. To avoid potential monotonic increasing or representation collapse~\cite{grill2020bootstrap}, we adopt a non-trainable target encoder (or momentum encoder) $g^{(m)}$ to compute the target next-state embedding $z_{t+1}^{(m)}$, which is updated by exponential moving average (EMA) of encoder $g$~\cite{he2020momentum}. The dynamics consistency loss of the transition is defined as
\begin{equation}\label{eq:dynloss}
\ell^{\text{dyn}}(\tau):= D^{\text{dyn}}(z_{t+1}, z_{t+1}^{(m)})
\end{equation}
where $D^{\text{dyn}}(\cdot, \cdot)$ is a similarity function between two inputs. Similar to prior work~\cite{schwarzer2020data, ye2021mastering}, we adopt SimSiam-style loss~\cite{chen2021exploring}. See Appendix~\ref{app:dyn_loss} for more details. 

Note that the dynamics consistency loss can be extended to longer transition sequence sampled from replay buffer: Given $\tau = (s_t, a_t, s_{t+1}, \dots, s_{t+K})$, we can iteratively get the prediction $z_{t+k+1} = h(z_{t+k},a_{t+k})$ as shown in fig.~\ref{fig:framework}. The final dynamics loss is $\ell^{\text{dyn}}(\tau) = \sum_{k=1}^{K}D^{\text{dyn}}(z_{t+k}, z_{t+k}^{(m)})$.

Although our method takes the consistency on transition dynamics into consideration, the main target is to capture the state structure relation in a latent embedding space instead of predicting the dynamics precisely as model-based RL~\cite{hafner2019learning, hafner2020mastering}, which requires a much larger number of parameters especially for tasks with high-dimensional state space and complex dynamics.

\subsection{Feasibility Consistency}

In safe RL, the agent performance is not only related to the dynamics estimation and reward critic learning, but also heavily depends on the constraint estimation. Both overestimation or underestimation of constraint violation can reduce the final reward: the formal makes the policy overly conservative while the latter can leads to higher Lagrangian multiplier $\lambda$ (i.e., a larger cost penalty coefficient) during learning. Therefore, to improve the safety awareness of the representation in safe RL, we propose to add a feasibility consistency loss in learning objective. 

We first define the feasibility score as maximum discounted cost: 
\begin{equation}
\label{equ: feasibility function}
 F^\pi(s,a) := \E_{\rho\sim\pi} \left[\max_t \gamma^t c(s_t,a_t)|s_0=s,a_0=a\right]. 
\end{equation}
The formulation of feasibility score is closely related to Hamilton-Jacobi (HJ) reachability in safe control theory and state-wise safe RL~\cite{bansal2017hamilton, fisac2019bridging, yu2022reachability}. The HJ-reachability-based methods use a much more informative safe signal to solve a safe RL problem with stricter constraint. Specifically, the reachability is computed based on a dense state constraint function (e.g., the distance to hazard) and indicates the level of state-wise safety. However, such dense cost may not be accessible in practical application. On the contrary, our method utilizes feasibility as an representation learning supervision instead of enforcing state-wise constraint satisfaction, which boosts the performance in general sparse-cost setting. 

The feasibility score satisfies the Bellman equation $F^\pi = \gP_F^{\pi}F^\pi$, where the corresponding Bellman operator is
\begin{equation}
\begin{aligned}
    &\gP_F^{\pi} F^\pi(s,a) := (1-\gamma) c(s,a)\\
    &\quad+ \gamma\max\left\{c(s,a), \E_{s'\sim\gT(\cdot|s,a), a'\sim\pi(\cdot|s')} F^\pi(s',a')\right\}.
\end{aligned}
\end{equation}
Different from previous~\cite{fisac2019bridging}, we adopt expectation over next action $\E_{a'}$ instead of maximization in Bellman operator, which is also commonly used in off-policy algorithms~\cite{lillicrap2015continuous, fujimoto2018addressing}.

The following proposition further shows that the feasibility score is also a safety indicator of future trajectory, which thus can be incorporated into safety-aware representation learning.

\begin{proposition}
\label{prop:1}
If the cost function $c$ is binary and the discount factor $\gamma\to 1$, then $(1-F^\pi(s,a))$ is equal to the probability of every following state-action is safe, i.e.,
\begin{equation}
    1-F^\pi(s,a) = \Pr\left(\bigcap_{(s_t,a_t) \sim \rho}\{c(s_t, a_t)=0\}\right),
\end{equation}
where $\rho$ is the trajectory starting with $(s,a)$ sampled by policy $\pi$.
\end{proposition} 

The proof is in Appendix~\ref{app:proof}. Proposition~\ref{prop:1} shows that the defined feasibility score is a indicator the level of safety. 

Therefore, we adopt feasibility score as a supervision signal to extract safety-related information. Specifically, given a sampled sub-trajectory $\tau=(s_t,a_t,r_t, c_t,s_{t+1},\dots)$, we apply a feasibility prediction head $\tilde{f}: \gZ\to\sR$ upon the learned embedding and minimize the loss between the predicted score and bootstrap estimate: 
\begin{equation}
    \ell^{\text{fea}}(\tau) = D^{\text{fea}}\left(\tilde{f}(z_t), F^{(m)}(s_{t})\right), 
\end{equation}
where $z_t=g(s_t)$, $F^{(m)}(s_{t}) = \max\{c_t, \gamma \tilde{f}(z_{t+1}^{(m)})\}$ \footnote{By definition, the target feasibility score is $\max\{c_t, (1-\gamma)c_t+\gamma \tilde{f}(z_{t+1}^{(m)})\}$, but we can ignore $(1-\gamma)c_t$ when cost function is binary.} denotes the bootstrap estimate of feasibility score for state $s_{t}$, and $D^{\text{fea}}(\cdot, \cdot)$ denotes the distance between two feasibility scores.

Similarly, for longer sequence, we can further extend the loss as: $\ell^{\text{fea}}(\tau) = \sum_{k=0}^{K-1}D^{\text{fea}}(\tilde{f}(z_{t+k}), F^{(m)}(s_{t+k}))$.
% F^{(m)}(s_{t+k})) = \max\{c_{t+k}, \gamma\tilde{f}(z_{t+k+1}^{(m)})\}

Since we predict the feasibility by a representation $z$ of state $s$ without the action, the true feasibility $F^\pi(s, \cdot)$ is actually a distribution when inputting different actions. When such distribution is widespread, it may cause large estimation error if we use a single expected value to represent. Therefore, we leverage the idea of distributional RL~\cite{bellemare2017distributional} and apply discrete regression~\cite{schrittwieser2020mastering, hafner2020mastering, schwarzer2020data} to learn the prediction head. Specifically, we discretize the output space into several buckets and compute the target discrete distribution $d_{F^{(m)}}(s_t)$ by projecting the target feasibility value $F^{(m)}(s_t)$ into buckets. Along with the predicted discrete distribution $d_{\tilde{f}}(z_t)$ by feasibility head $\tilde{f}$, the optimization loss is defined as the KL divergence: 
\begin{equation}
    D^{\text{fea}}\left(\tilde{f}(z_t), F^{(m)}(s_{t})\right) = \KL\left(d_{F^{(m)}}(s_t) \| d_{\tilde{f}}(z_t)\right).
\end{equation}
We also adopt the discrete regression in baselines (e.g., to predict cost value $V_c$ or one-step cost) for fair comparison.

\subsection{Summary of the Proposed Method}

% The dynamics consistency loss affects the encoder $g$ and transition model $h$ while the feasibility consistency loss affects the encoder $g$ and feasibility prediction head $\tilde{f}$. 
Let $\theta$ denote the parameters of $\{g, h, \tilde{f}\}$, the final objective for representation learning is 
\begin{equation}
\label{eq:learn_obj}
    \gL_\theta = \E_{\tau\sim \gD} \left[\gL_\theta^{\text{dyn}} (\tau) + \lambda^{\text{fea}} \gL_\theta^{\text{fea}}(\tau)\right].
\end{equation}

Since the update of state representation may cause large noise in input space when optimizing RL objective (e.g., both actor and critic loss), we take the target representation $z^{(m)}$ generated by target encoder as their inputs when training the policy and value function of the agent. 

By incorporating our representation learning method into an on-policy or off-policy model-free safe RL algorithm, we obtain \textbf{F}easibility \textbf{C}onsisent \textbf{S}afe \textbf{RL} (FCSRL). The main procedure of FCSRL is summarized in Algorithm \ref{alg:fcsrl}. 

When using an on-policy safe RL algoithm (e.g., PPO-Lagrangian) as the base RL method, we replace the bootstrapped value and feasibility score by Monte Carlo estimation from sampled on-policy trajectories: $V_c^{(m)}(s_i) = \sum_{t=i}^T \gamma^t c_t, F^{(m)}(s_i) = \max_{i\leq t\leq T} \gamma^t c_t$, which aligns the objective of representation learning with the critic learning in on-policy RL. 

\begin{algorithm}[htb]
\caption{Feasibility Consistent Safe RL}
\label{alg:fcsrl}
% {\bfseries Input:} \raggedright rollouts $T$
\raggedright Initialize policy $\pi$, value functions, Lagrangian multiplier; \\
Initialize encoder and target encoder $g, g^{(m)}$, transition model $h$ and feasibility head $\tilde{f}$, which are parameterized by $\theta_{g}, \theta_{gm}, \theta_h, \theta_{f}$ respectively; $\theta_{gm} \leftarrow \theta_g$. \\
% Denote $\theta=\{\theta_{g}, \theta_h, \theta_{f}\}$ as the set of trainable parameters; \\
Initialize replay buffer $\gD$. \par
\begin{algorithmic}[1] % The number tells where the line numbering should start
\WHILE{Training}
\STATE Collect experience $\{(s,a,r,c,s')\}$ by policy $\pi$ and add it to replay $\gD$;
\STATE Sample a minibatch of length-$K$ sub-trajectories $B$ from $\gD$;
\STATE $\triangleright$ \textit{state representation learning}
\STATE Compute the $\gL_\theta$ by Eq.(\ref{eq:learn_obj}) by sub-trajectories $B$.
\STATE Update the $g, h, \tilde{f}$ by loss $\gL_\theta$;
\STATE Update $\theta_{gm}$ by EMA of $\theta_g$;
\STATE Compute the target representation $z^{(m)}=g^{(m)}(s)$;
\STATE $\triangleright$ \textit{safe reinforcement learning}
\STATE Update value functions with $z^{(m)}$ as input by $B$;
\STATE Update policy with $z^{(m)}$ as input by $B$;
\IF{Lagrangian flag}{
\STATE Update Lagrangian multiplier.
}
\ENDIF
\ENDWHILE
% \STATE $\triangleright$ \textit{policy extraction}
% \FOR{training iteration $j$}
% % \STATE $\triangleright$ \textit{learn policy}
% \STATE Update policy $\pi_\theta$ by Eq.(\ref{eq: policy learning}).
% \ENDFOR
\end{algorithmic}
\end{algorithm}
\vspace{-3mm}

\subsection{Comparison with Value Consistency}
In previous work, another objective for state representation learning is value consistency~\cite{schrittwieser2020mastering, ye2021mastering, hessel2021muesli, farquhar2021self}, which enforces the learned state representation to be able to predict the corresponding value function and has proven to improve the reward performances in standard RL setting. However, in safe RL problems, the consistency objective on cost value does not exhibit similar advantages due to sparsity of the cost signals, which results in large in-continuity and roughness (non-smoothness) of the value estimation, and thus increasing the difficulty of prediction. 

On the contrary, within our method, the proposed feasibility function is always \textit{smoother} than the original cost value function, which is introduced in the following proposition.

\begin{definition}[Temporal smoothness] 
    Given a trajectory $\rho=\{s_0,a_1,s_1,\dots, s_T\}$, with a slight abuse of notation, we denote an arbitrary state-action function $f(s_t, a_t)$ based on $\rho$ as $f(t)$. 
    % where $t$ is the time step in trajectory $\rho$ and $s_t, a_t$ is the corresponding state-action pair. 
    The temporal smoothness of $f$ along the trajectory $\rho$ is defined as: 
    \begin{equation}
        L(f, \rho) := \E_{\rho} |f(t) - f(t+1)|,
    \end{equation}
    where the expectation is with respect to $t$ along the given trajectory $\rho$. The smoothness $L$ characterizes the rate of change of the function along the specified trajectory. % , and the smaller, the smoother
\end{definition}
\begin{proposition}
\label{proposition: smoothness}
    The single trajectory estimation of feasibility score is temporally smoother than cost value function for any trajectory, i.e.,
    \begin{equation}
        L(\hat{F}, \rho) \leq L(\hat{V}_c, \rho), \quad \forall \rho,
    \end{equation}
where given $(s_t, a_t)\sim\rho$, the functions $\hat{F}$ and $\hat{V}_c$ is defined as:
\begin{equation}
\begin{cases}
    \hat{F}(s_t, a_t) =  \max_{i \geq t} \gamma^{t-i} c(s_i,a_i) \\
    \hat{V}_c(s_t, a_t) =  \sum_{i \geq t} \gamma^{t-i} c(s_i,a_i)
\end{cases}.
\end{equation}
\end{proposition}
% \begin{equation}
%     \hat{F}(s_t, a_t) =  \max_{i \geq t} \gamma^t c(s_i,a_i), \hat{V}_c(s_t, a_t) =  \sum_{i \geq t} \gamma^t c(s_i,a_i),
% \end{equation}
The proof is available in Appendix~\ref{app: proof 2}.

\begin{figure}[h]
    \centering
    \begin{subfigure}[b]{0.49\columnwidth}
         \centering
         \includegraphics[width=0.9\textwidth]{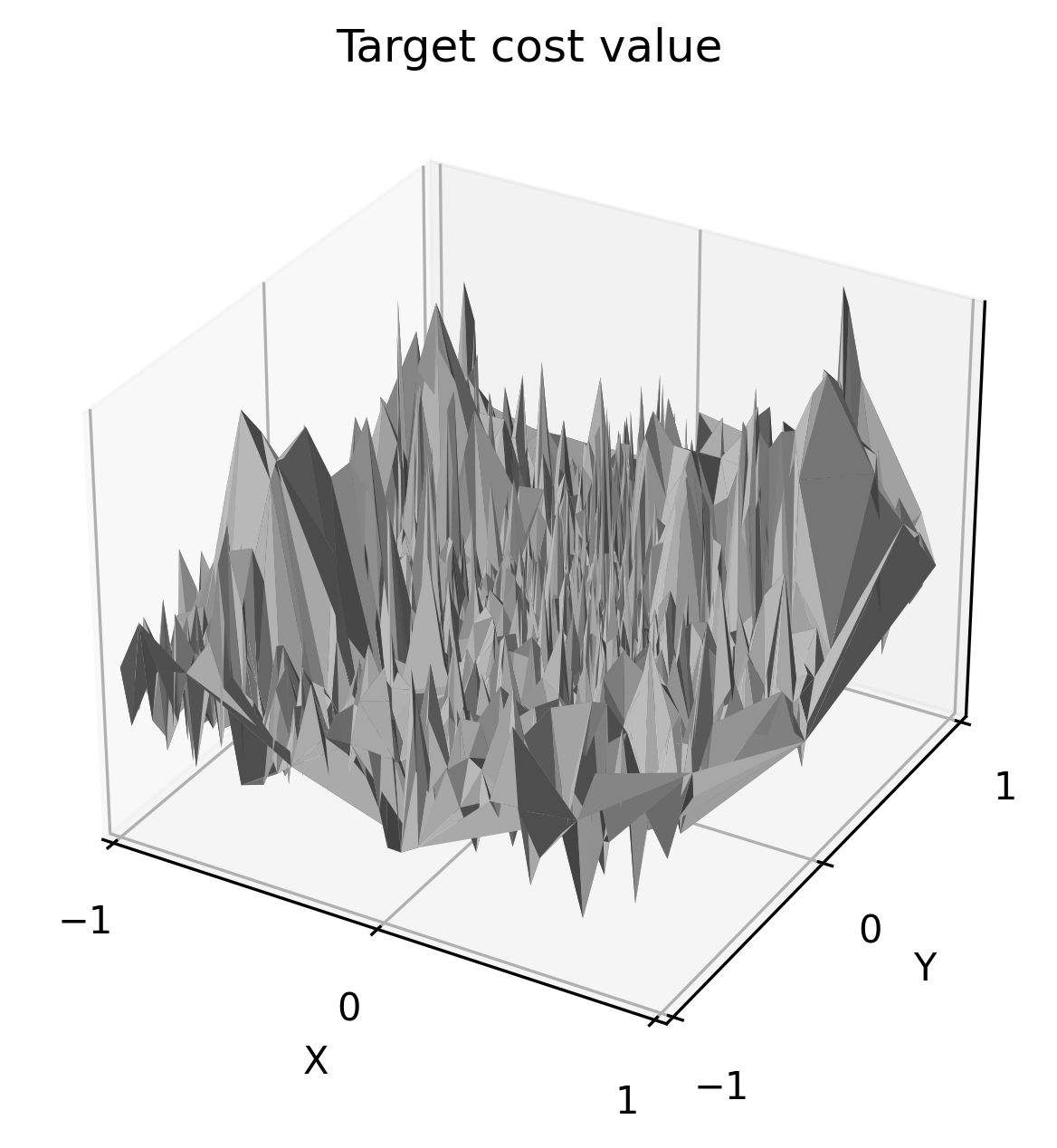}
         % \caption{Base method: TD3-Lagrangian.}
         % \label{fig:hm1}
     \end{subfigure}
     \hfill
     \begin{subfigure}[b]{0.49\columnwidth}
         \centering
         \includegraphics[width=0.9\textwidth]{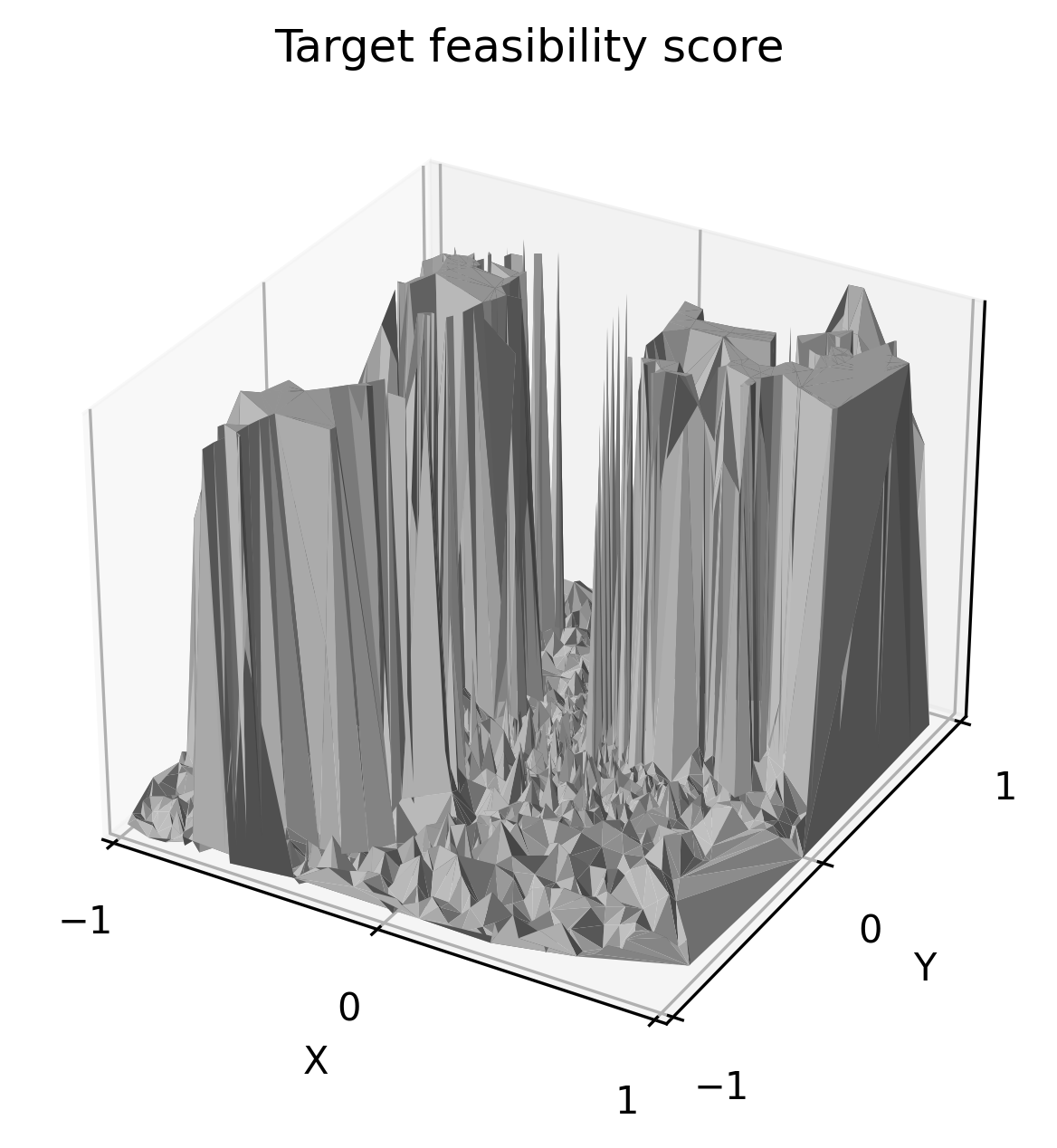}
         % \caption{Base method: TD3-Lagrangian.}
         % \label{fig:hm1}
     \end{subfigure}
     \caption{The landscape of target cost value $V_c^{(m)}(s)$ and target feasibility score $F^{(m)}(s)$ obtained by bootstrap estimation in PointGoal2 task. The X-Y axis means the \textit{coordinate of agent} when its state is $s$. The values of z-axis have been rescaled. See Appendix~\ref{app:landscape} for more details.}
     \label{fig:landscape}
     \vspace{-3mm}
\end{figure}

Furthermore, to empirically compare the cost value and feasibility score, we visualize the target estimations of them in the same region of PointGoal2 task, which are computed by Bellman bootstrap. The figure~\ref{fig:landscape} shows that the feasibility score is much less noised than the value function and matches better with unsafe region. Therefore, feasibility score is a better safety-awareness signal for representation learning in sparse cost setting. 

\section{Experiment}
In the experiment part, we empirically test the performances of proposed representation method. Particularly, we focus on two main questions: (1) Does FCSRL exhibit a consistent advantage over baselines across a wide range of environments, both in vector-state and image-based settings? (2) Is FCSRL capable of learning a safety-aware state representation to boost the performance of safe RL algorithm? 
% \TODO{proof read, remove some introduction of baselines}

\subsection{Tasks}
To answer the above questions, we use $6$ vector-state and $3$ image-based continuous robotic control tasks as our testbeds adopted from safety-gymnasium~\cite{ji2023safety}, a widely used evaluation benchmark by previous work for safe RL~\cite{liu2023datasets}. 

\subsection{Results on Vector-state Tasks}

We first compare our method with previous representation learning baselines in low-dimensional state-input tasks.

\subsubsection{Baselines}
% We use $(s_t,a_t,r_t,c_t,s_{t+1})$ to denote a transition and $\{s_t, a_t, r_t,c_t, \dots, s_{t+K}\}$ to denote a sub-trajectory of states, actions, rewards, and costs. $z_{t+k}=g(s_{t+k})$ denotes the representation for each state. 

For ease of reading, we omit the expectation over transition or sub-trajectory in the learning objective when introducing the baselines. We adopt following baselines:

\textbf{Raw state input}: Without encoding state into embedding, both policy and value function take the raw state as input.

\textbf{Forward raw model} predicts the next state $s_{t+1}$ based on the embedding $z_t$ and $a_t$. Specifically, we train an additional state prediction model $f:\gZ\times\gA\to \Delta(\gS)$ which outputs a distribution over next state. The learning objective is $\min -\log P(s_{t+1}|f(z_t,a_t))$.

\textbf{Forward latent model} predicts the next state $z_{t+1}$ based on the embedding $z_t$ and $a_t$. The predictor $f:\gZ\times\gA\to \Delta(\gZ)$ outputs the distribution over next state representation. The learning objective is $\min -\log P(z_{t+1}|f(z_t,a_t))$, which can be generalized to longer sequence as $\min -\sum_{k=0}^{K-1}\log P(z_{t+k+1}|f(\hat{z}_{t+k},a_{t+k}))$, where $\hat{z}_{t+k}$ is the mean output of the last prediction.

\textbf{Inverse model} predicts the action $a_t$ based on the embeddings $z_t, z_{t+1}$. The predictor $f:\gZ\times\gZ\to \Delta(\gA)$ outputs the distribution over action and the learning objective is $\min -\log P(a_{t}|f(z_t,a_{t+1}))$.

% \textbf{Value equivalence (VE) model}~\cite{grimm2020value}. Similar to above value consistent model, VE model enforces the learned representation to be predictive of value. However, to obtain the equivalent value to real rollouts, VE model learns the critic in an end-to-end manner instead of learning an additional value prediction head like value consistent model. VE model directly feeds the $z_t$ encoded by encoder $g$ to critic, and jointly use critic loss and other representation loss to train encoder, which. The learning objective is
% \begin{equation}
%     \min -\left(\gL^{\text{dyn}} + \lambda\text{MSE}(V_c(z_{t}), V_c^{(m)}(z_{t}))\right)
% \end{equation}
% where $V_c^{(m)}(z_{t}) = c_t + \gamma V_c(z_{t+1})$ is a bootstrap estimate of value at time $t+1$ and is gradient-stopped as in many RL algorithms~\cite{haarnoja2018soft, fujimoto2018addressing}.

\textbf{Temporal contrastive learning (TCL)} applies a contrastive loss between $z_{t}$ and $z_{t+1}$. We use $\exp(z_1^T W z_2)$ as similarity function, where $W$ is a trainable square matrix, and use InfoNCE loss~\cite{oord2018representation} as learning objective: 
\begin{equation}
    \min -f(z_{t}^T W z_{t+1}) + \log \sum_{\tilde{z}\in Z} \exp(z_{t}^T W \tilde{z})
\end{equation}
where $Z=\{z_{t+1}^{(1)}, z_{t+1}^{(2)}, \cdots, z_{t+1}^{(N)}\}$ is a batch of embeddings randomly sampled from replay buffer. It can also be generalized to long sequence with multiple weight matrices $\{W_1, \dots, W_k\}$.
% \begin{equation}
%     \min -\sum_{k=0}^{K-1}\log \frac{\exp(z_{t}^T W_k z_{t+k+1})}{\sum_{\tilde{z}} \exp(z_{t}^T W_k \tilde{z})}
% \end{equation}
We also use target encoder $g^{(m)}$ (non-trainable and updated by EMA of online encoder $g$) to obtain the representation of next state $z_{t+1}$, which empirically improves the reward performance in prior work~\cite{yang2021representation}.

\textbf{SALE}~\cite{fujimoto2023sale} learns an encoder $g$ and transition model $h$. The learning objective is to minimize the MSE between $z_{t+1}$ and $\hat{z}_{t+1}=h(z_t, a_t)$. SALE also rescales the embedding $z$ by average L1 normalization to prevent the representation collapse. Meanwhile, SALE finds that inputting both state and its representation into policy and value function in RL empirically improves the final performance in vector state tasks.

\textbf{Value consistent model (VC)}: The main idea of value consistent model is to enforce the learned embedding to predict cost value function with a prediction head $\tilde{v}:\gZ\to\sR$, i.e., $\tilde{v}(z_t) = V_c(s_t), z=g(s_t)$. Meanwhile, we observe that the value equivalence~\cite{grimm2020value} may deteriorate the final performance, which learns the critic in an end-to-end manner by 1) directly feeding the $z_t$ to RL critic, and 2) jointly using critic loss and other representation loss to train encoder $g$. This phenomenon has also been observed by previous work~\cite{farquhar2021self}. Therefore, we still learn the representation with an additional value prediction head. The only difference from FCSRL is that the VC models predicts the cost value and we keep all other settings the same, e.g., using target encoder and discrete regression loss.

% \textbf{Value equivalence (VE) model}~\cite{grimm2020value}. Similar to above value consistent model, VE model enforces the learned representation to be predictive of value. However, to obtain the equivalent value to real rollouts, VE model learns the critic in an end-to-end manner instead of learning an additional value prediction head like value consistent model. VE model directly feeds the $z_t$ encoded by encoder $g$ to critic, and jointly use critic loss and other representation loss to train encoder, which. The learning objective is

For fair comparison, we adopt the same network architecture of neural network. Specifically, we parameterize the encoder $g$, transition model $h$, and all prediction heads as two-layer MLPs. Following SALE, FCSRL and VC also input the concatenation of state and representation to the RL policy and value function.

\begin{figure*}[htb]
    \centering
    \begin{subfigure}[b]{0.95\textwidth}
     \centering
     \includegraphics[width=1.0\textwidth]{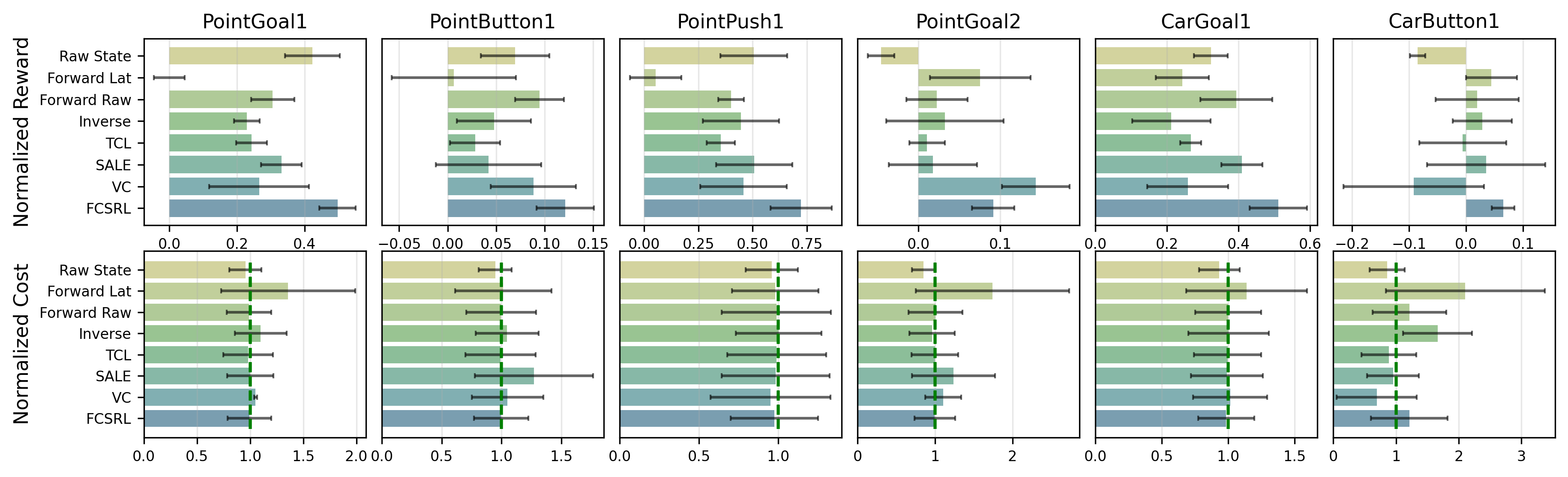}
     % \caption{Base RL algorithm: PPO-Lagrangian.}
     % \label{fig:hm1}
    \end{subfigure}
    
    \begin{subfigure}[b]{0.95\textwidth}
         \centering
         \includegraphics[width=1.0\textwidth]{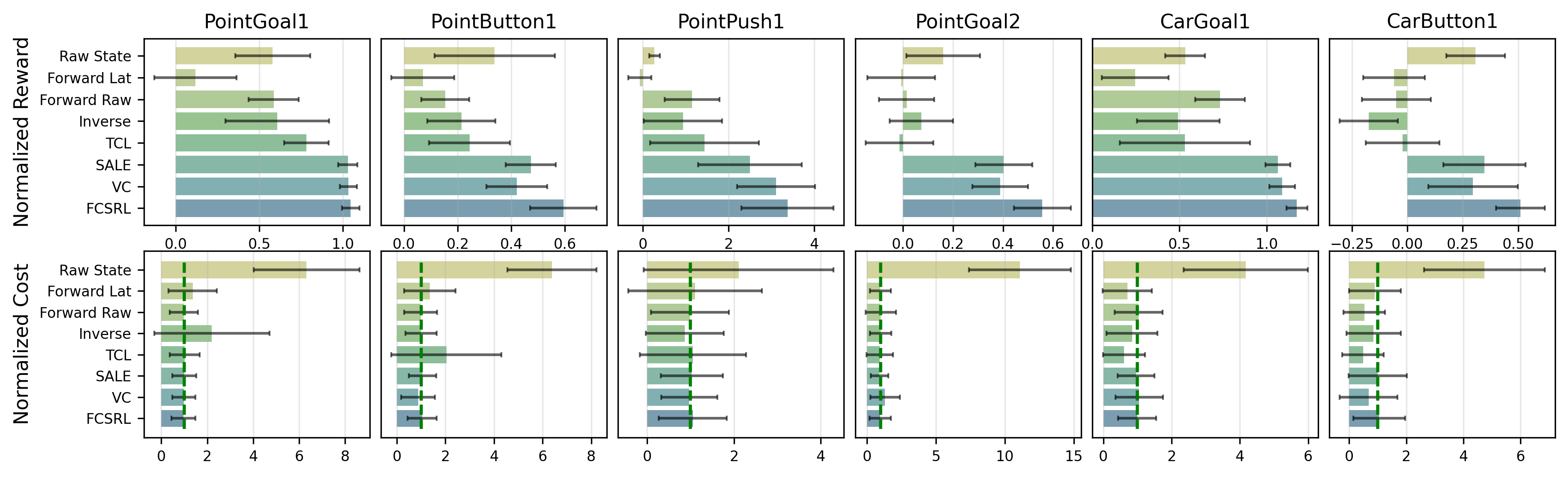}
         % \caption{Base RL algorithm: TD3-Lagrangian.}
         % \label{fig:hm1}
     \end{subfigure}
     \caption{The converged performances of different representation learning methods based on PPO-Lagrangian (\textbf{top}) and TD3-Lagrangian (\textbf{bottom}). The error bar indicates the standard deviation of 5 seeds. The \textit{green dash line} in normalized cost plots indicates the constraint threshold.}
     \label{fig:main_result}
     \vspace{-3mm}
\end{figure*}

\subsubsection{Evaluation Results}

To compare the performances of our method with above representation learning baselines, we test them with two different base safe RL algorithms: an on-policy method \texttt{PPO-Lag}, and an off-policy one \texttt{TD3-Lag}, which augment PPO~\cite{schulman2017proximal} and TD3~\cite{fujimoto2018addressing} with Lagrangian method~\cite{ray2019benchmarking}. We also use PID Lagrangian update~\cite{stooke2020responsive} to improve the stability of cost performance. We adopt cost limit $\epsilon=10$ for \texttt{TD3-Lag} experiments and $\epsilon=25$ for \texttt{PPO-Lag} because almost all representation learning methods with \texttt{PPO-Lag} fail to get a positive reward while satisfying constraint when $\epsilon$ is small. We set prediction length $K=4$ for forward latent, TCL, SALE, VC and FSCRL. We train every method for 2M environment steps. The training curve is attached in Appendix~\ref{app:trainig_curve}. 

In fig.~\ref{fig:main_result}, we report the normalized rewards and costs of baselines and our method. The normalized reward is computed by NR $= (R-R_l) / (R_h - R_l)$, where $R$ is the unnormalized reward return and $R_l, R_h$ denote the reward performances of \textit{random policy} and \textit{unconstrained PPO policy} respectively on the given task. The normalized cost is computed by NC $= C / \epsilon$, where $C, \epsilon$ are the unnormalized cost return and cost threshold.

% \TODO{analysis}
% 1. Overall, FCSRL exceeds all baselines on most tasks while satisfying the constraint.
% 1. TD3 better than PPO; but TD3 with raw state input cannot satisfy the constraint. 
% 2. 
As illustrated in fig.~\ref{fig:main_result}, the cost performances of most representation learning methods converge to the preset cost limit, which validates that the Lagrangian method is able to balance the reward and cost performances and tunes the value of Lagrangian multiplier $\lambda$ according to the discrepancy between the cost performance and the constraint threshold.

Overall, the methods based on \texttt{TD3-Lag} outperform those based on \texttt{PPO-Lag} but the raw-state-input \texttt{TD3-Lag} fails to meet the constraint criteria. FCSRL exceeds all baselines on most tasks with either \texttt{PPO-Lag} or \texttt{TD3-Lag}. In Goal1 tasks, the advantage margin of FCSRL over SALE and VC model seems to be relatively small. This may be attributed to their close performances to the theoretical maximum reward and one evidence is that their rewards exceed unconstrained PPO (with normalize reward = $1$). Additionally, FCSRL achieves superior performance on Push1 task than unconstrained PPO, which validates the advantages of feasibility consistency in representation learning. 

Besides, the SALE, VC, and FCSRL have relatively higher performances than the remaining baseline, suggesting the advantages in feeding both state and representation for low-dimensional state tasks. It may stem from reward-related information loss (e.g., the distance of agent to "Goal" is closely related to the reward) in representation learning, although the learned embedding captures the safety-related information in state space. We also provide a comparison in Sec.\ref{sec:ablation}. Meanwhile, it is noteworthy that the forward raw method (which predicts raw state) outperforms forward latent (which only predicts latent embedding) in most tasks, which further highlights the necessity of retaining the information in raw state.

Compared to VC model, the reward margin of FCSRL over value consistent model is larger on tasks where constraint is harder to satisfy (this means the reward is much smaller than the unconstrained tasks, e.g., PointButton1, PointGoal2 and CarButton1), which shows the advantages of using feasibility score to supervise the embedding learning. Furthermore, We provide a detailed comparison of learned embedding $z$ by two methods in Appendix~\ref{app:compare_vc_fc}. 

\subsection{Results on Image-based Tasks}

We further consider high-dimensional image-based tasks, where the image-input RL usually encounters the issue of data inefficiency and it requires an encoder to extract a low-dimensional representation from high-dimensional raw observation. Therefore, we compare our method with other vision-based representation learning methods to verify the effectiveness of feasibility consistency.

\begin{figure}[htb]
    \centering
    \includegraphics[width=0.98\linewidth]{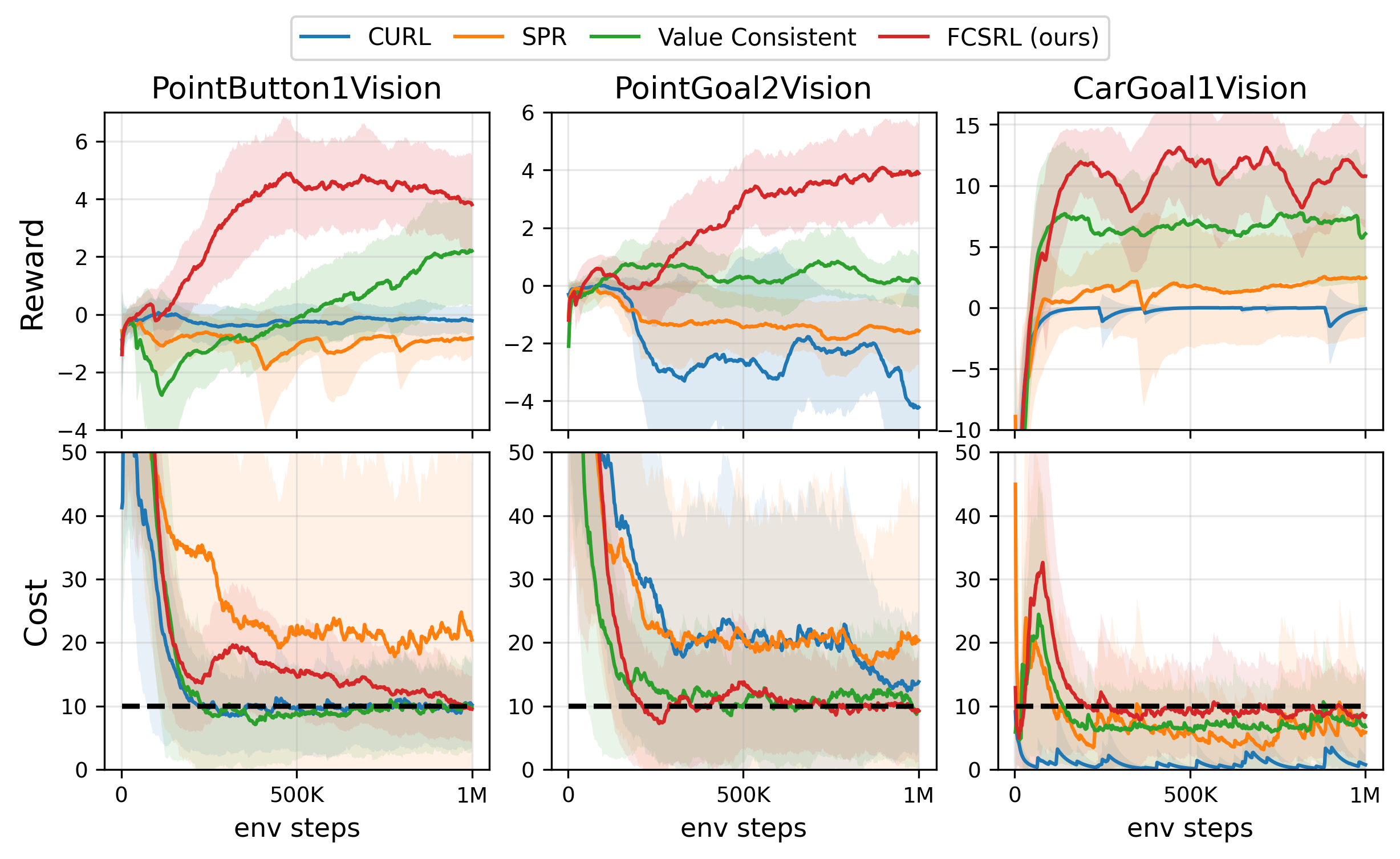}
    \caption{Training curve of image-based tasks. The black dash line is the cost limit. The shadow region is the standard deviation of 5 seeds.}
    \label{fig:image-based_train_curve}
    \vspace{-3mm}
\end{figure}

We adopt following image-based representation learning baseline: 1) \textbf{CURL}~\cite{laskin2020curl} extracts the embedding by applying contrastive learning on two augmentation given the same original image; 2) \textbf{SPR}~\cite{schwarzer2020data} learns the representations by applying contrastive learning on adjacent states in the same trajectory; and 3) \textbf{Value consistent model}. We directly input the raw image to encoder without augmentation for SPR, VC and FCSRL. To exclude the influence of network structure, we keep it the same for fair comparison. We use \texttt{TD3-Lag} as the base safe RL algorithm, where the policy and value function take the low-dimensional embedding $z$ as input. We report the training curve in fig.~\ref{fig:image-based_train_curve}.

The results clearly demonstrate that our method significantly surpasses the baseline models. Specifically, CURL fails to achieve high rewards and exhibits excessive conservatism in the CarGoal1 task, with both reward and cost metrics nearing zero. This limitation arises because CURL only learns an embedding for a single state, focusing solely on vision representation and neglecting the temporal features crucial for sequential decision-making. SPR, on the other hand, does not meet the constraints in the PointButton1 and PointGoal2 tasks. This leads to an increase in the Lagrangian multiplier, consequently diminishing the reward performance during training. In contrast, both the value-consistent model and FCSRL excel in terms of reward and cost, underscoring the benefits of steering representation learning with cost-related signals. Furthermore, the performance of FCSRL indicates that representations with feasibility consistency are more effectively learned, exhibiting a better capability to distill a low-dimensional embedding from high-dimensional image states.

\subsection{Performances with Different Cost Limits}
In this section, we aim to study how the safety-awareness of learned representation affects the safe RL performance. Based on the intuition that safe RL policy performance increasingly depends on safety-aware representations as cost constraints become more stringent, we evaluate the performance discrepancies among different representation learning methods under cost limit $\epsilon\in \{5,10,20,30,40\}$.

\begin{figure}[htb]
    \centering
    \includegraphics[width=0.8\linewidth]{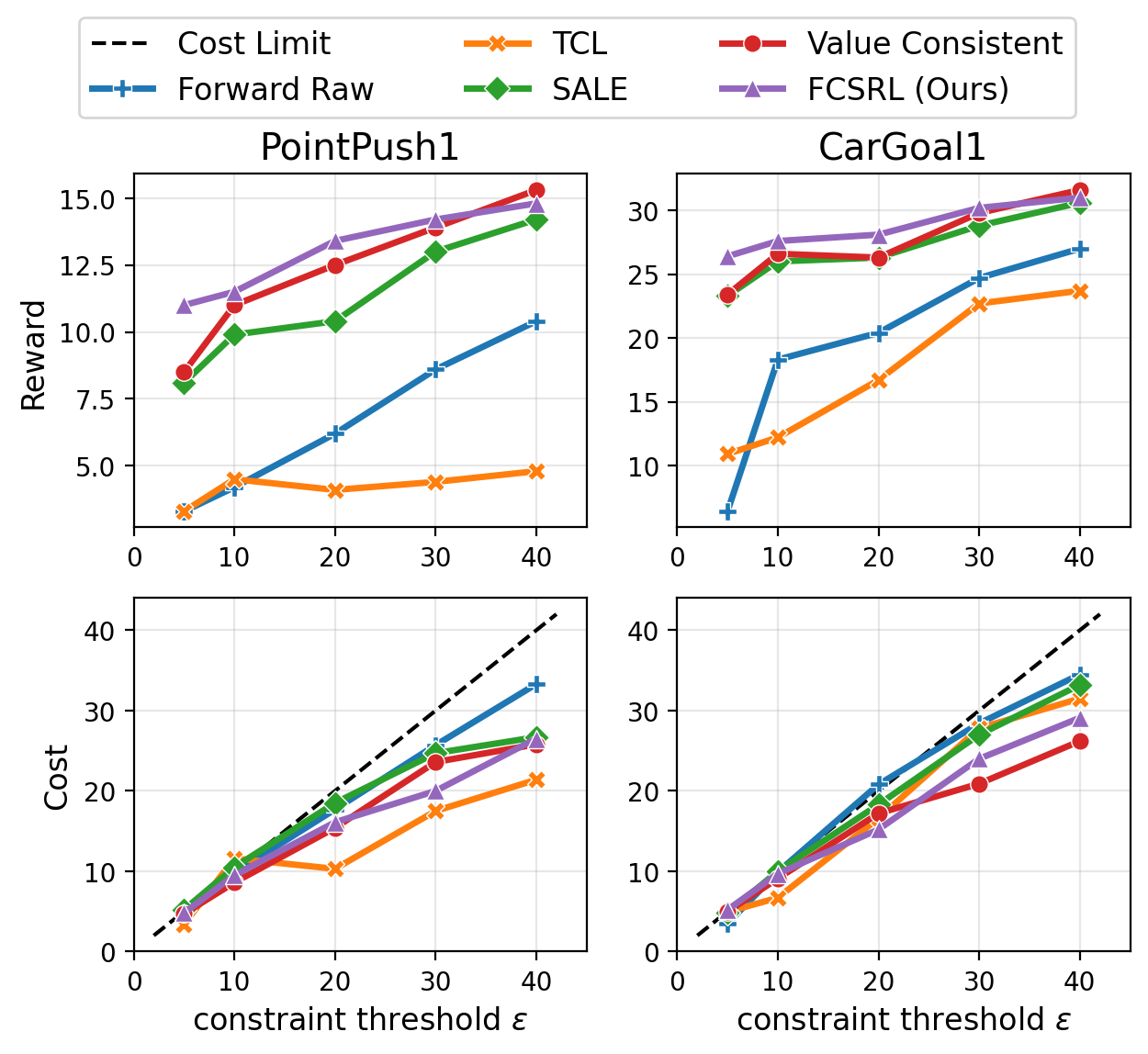}
    \caption{Comparison of reward and cost performances with different constraint thresholds.}
    \label{fig:varied_limit}
\end{figure}

We test the performances of forward raw, TCL, SALE, VC, and FCSRL based on \texttt{TD3-Lag} on PointPush1 and CarGoal1 tasks. The converged reward and cost are plotted in fig.~\ref{fig:varied_limit}. Overall, most representation methods have a monotonic reward increasing when we set larger cost limit. When cost limit $\geq 20$, the final cost performances do not exactly converge to the given threshold. This is because the corresponding unconstrained policy (i.e., reward maximization without considering cost) is already constraint satisfactory. 

The FCSRL outperforms the other baselines, which suggests a better safety-related feature extraction by our method. Regarding constraint strength, FCSRL achieves very similar performance to SALE and VC models when cost $>30$; however, the performance advantage of FCSRL grows, displaying the boost on safe RL from the safety-aware representation. This not only empirically validates the aforementioned intuition but also highlights the effectiveness of our method under stricter safety requirements.

\subsection{Ablation Studies}
\label{sec:ablation}
\textbf{The effectiveness of each components}. We test it by remove each component in representation learning. We test each variant based on \texttt{TD3-Lag} on PointGoal2 and PointButton1 tasks and set cost limit as 10. The results are reported in table~\ref{tab:ablation}.

\begin{table}[htb]
\centering
\caption{Ablation study on each component in FCSRL.}
\label{tab:ablation}
\resizebox{0.95\linewidth}{!}{
\begin{tabular}{@{}lcccc@{}}
\toprule
                      & \multicolumn{2}{c}{PointGoal2} & \multicolumn{2}{c}{PointButton1} \\ \midrule
                      & reward         & cost          & reward          & Cost           \\ \midrule
full FCSRL            & 12.7±2.4       & 10.0±7.8      & 10.7±2.3        & 9.8±6.0        \\
w.o. dynamics loss    & 10.3±2.9       & 9.4±8.5       & 10.5±2.6        & 9.6±5.5        \\
w.o. feasibility loss & 7.1±2.7        & 9.8±9.4       & 9.0±2.9         & 10.2±5.8       \\
only input $z$        & 5.9±3.3        & 9.7±0.8       & 7.3±4.3         & 9.6±6.3        \\ \bottomrule
\end{tabular}
}
\vspace{-3mm}
\end{table}

We observe that feasibility loss plays an importance role in representation learning of FCSRL. The performance of FCSRL has a large drop when only inputting $z$ but it is less affected by removing dynamics loss. This may because the FCSRL heavily relies on raw state capture to capture the features of dynamics in vector-state tasks if we input both state and its representation. Meanwhile, the raw state may contain additional information for reward-maximization optimization in RL, e.g., the features related to the reward but neglected by representation learning, which has also been observed in standard RL setting~\cite{fujimoto2023sale}. 
% It is also worth noting that FCSRL still has higher reward performance than best baseline (inverse model with 2.9 reward on PointGoal2 and 4.8 on PointPush1) even in the case of only inputting $z$. 

% \textbf{Choice of prediction signal}
% 2. ablation 2: change FC to predict value/cost

\textbf{Prediction length}.
Previous works observe that increasing prediction length facilitates the representation learning~\cite{schwarzer2020data, yang2021representation}. Therefore, we further present a comparison with different prediction lengths on vector-state PointPush1 task. We still use \texttt{TD3-Lag} as base RL algorithm and set cost limit as 10.

\begin{table}[htbp]
\centering
\caption{The performances with different prediction lengths.}
\label{tab:length}
\resizebox{0.95\linewidth}{!}{
\begin{tabular}{@{}lccccc@{}}
\toprule
\multicolumn{2}{l}{Prediction length $K$}                                          & 2        & 4         & 6         & 8         \\ \midrule
\multirow{2}{*}{\begin{tabular}[c]{@{}l@{}}Forward\\ latent\end{tabular}} & reward & -1.1±1.3 & -0.6±0.9  & -0.4±0.8  & -0.5±1.0  \\
& cost   & 8.9±11.5 & 10.1±13.5 & 8.8±10.0  & 10.1±16.7 \\ \midrule
\multirow{2}{*}{TCL}                                                      & reward & 2.2±3.6  & 4.9±1.6   & 4.7±2.1   & 5.1±0.6  \\
& cost   & 9.3±9.2  & 8.4±8.0   & 10.2±15.8 & 9.5±17.3  \\ \midrule
\multirow{2}{*}{SALE}                                                     & reward & 7.7±4.4  & 8.6±2.4   & 8.6±3.0   & 9.1±2.6   \\
 & cost   & 9.4±7.4  & 10.5±7.1  & 10.5±7.1  & 10.5±7.3  \\ \midrule
\multirow{2}{*}{FCSRL}                                                    & reward & \textbf{9.7±2.5}  & \textbf{11.6±4.3}  & \textbf{12.4±3.4}  & \textbf{12.6±3.8}  \\
& cost   & \textbf{10.7±8.9} & \textbf{9.6±6.8}   & \textbf{9.6±8.0}   & \textbf{9.3±7.3}   \\ \bottomrule
\end{tabular}
}
\end{table}

The results in table~\ref{tab:length} show that most representation learning benefits as the prediction length extends, with our approach reliably surpassing the baseline methods across various lengths. However, we notice that the reward increase becomes less pronounced when $K\geq 4$. This trend suggests that the benefits of predicting longer sequences diminish for the actor and critic components of reinforcement learning.
% \vspace{-3mm}
\section{Conclusion}
In this paper, we present a novel framework for safe RL that leverages feasibility consistent representation learning to improve safety and efficiency. Through extensive experiments, we demonstrate that our approach outperforms existing methods by effectively balancing task performance and safety constraints. 
The capability of our model to extract safety-related features from complex environments and its application across various tasks underscore its potential to promote existing safe RL methods. One future direction of our work is to employ feasibility score as auxiliary signal for policy learning to achieve state-wise safety in safe RL.
% This work not only presents a novel framework for safety-aware representation learning but also sets the stage for future research aimed at further refining and expanding the applicability of safe policy learning.

\section*{Acknowledgements}
The research is partly supported by the National Science Foundation under grants CNS-2047454.

\section*{Impact Statement}
This paper presents work whose goal is to advance the field of Machine Learning. There are many potential societal consequences of our work, none which we feel must be specifically highlighted here.

\bibliography{ref}
\bibliographystyle{icml2024}

%%%%%%%%%%%%%%%%%%%%%%%%%%%%%%%%%%%%%%%%%%%%%%%%%%%%%%%%%%%%%%%%%%%%%%%%%%%%%%%
%%%%%%%%%%%%%%%%%%%%%%%%%%%%%%%%%%%%%%%%%%%%%%%%%%%%%%%%%%%%%%%%%%%%%%%%%%%%%%%
% APPENDIX
%%%%%%%%%%%%%%%%%%%%%%%%%%%%%%%%%%%%%%%%%%%%%%%%%%%%%%%%%%%%%%%%%%%%%%%%%%%%%%%
%%%%%%%%%%%%%%%%%%%%%%%%%%%%%%%%%%%%%%%%%%%%%%%%%%%%%%%%%%%%%%%%%%%%%%%%%%%%%%%
\newpage
\appendix
\onecolumn
\section{Proof for Theoretical Analysis}

\subsection{Proof of Proposition~\ref{prop:1}}
\begin{proof}

\label{app:proof}
Given a trajectory $\rho = \{s_1, a_1, ..., s_T, a_T\}$ with length $T$, by definition, we have
\begin{equation}
\label{equ: proof 1}
\max _t\left(c\left(s_t, a_t\right)\right)=1-\textbf{1}\left(\bigcap_{t=1}^{T} c\left(s_t, a_t\right)>0\right),
\end{equation}
which means along the trajectory $\rho$, the maximal value of $c\left(s_t, a_t\right)$ is the complementary of the event that at least one of the state-action pair $(s_t, a_t)$ violates the safety constraint. Then add expectations on both sides of (\ref{equ: proof 1}), we can get:
\begin{equation}
\begin{aligned}
    \E_{\rho\sim\pi}
\max _t\left(c\left(s_t, a_t\right)\right)= & 
1- \E_{\rho\sim\pi} \textbf{1}\left(\bigcap_{t=1}^{T} c\left(s_t, a_t\right)>0\right) \\
= &  1- \Pr\left(\bigcap_{(s_t,a_t) \sim \rho}\{c(s_t, a_t)=0\}\right)
\end{aligned}
\end{equation}
Since with definition (\ref{equ: feasibility function}), the left side is the feasibility function. Then we can conclude that:
\begin{equation}
    1-F^\pi(s,a) = \Pr\left(\bigcap_{(s_t,a_t) \sim \rho}\{c(s_t, a_t)=0\}\right)
\end{equation}
    
\end{proof}

\subsection{Proof of Proposition~\ref{proposition: smoothness}}
\label{app: proof 2}
\begin{proof}
    Given trajectory $\rho$ with length $T$, we first define functions $\hat{F}(i)$ and $\hat{V}(i)$:
    \begin{equation}
        \hat{F}(i) := \max_{t\geq i} \gamma^{t-i} c(s_t,a_t), \quad \hat{V}(i) := \sum_{t=i}^{T}\gamma^{t-i} c(s_t,a_t),
    \end{equation}
    where $i$ means time step. Then we can first observe that:
    \begin{equation}
        \hat{F}(i) \leq \hat{V}(i),\quad \forall \ i;
    \end{equation}
    Then we come to the temporal smoothness part. The smoothness is discussed with the following conditions:
    \begin{enumerate}
        \item If $c(i) = 0, \ \forall i \leq T$: Then $\sum \hat{F}(i) = \sum \hat{V}_c(i) = 0$. In this case, $L(\hat{F}, \rho) = L(\hat{V}_c, \rho) = 0$.
        \item If exist one unique $i_0$ such that $c(i_0) = 0$, then for the time step $i > i_0$, $\sum_{i>i_0} \hat{F}(i) = \sum_{i>i_0}  \hat{V}_c(i) = 0$ based on the analysis above. For the time step $i \leq i_0$:
        \begin{equation}
            \hat{F}(i) = 1, \quad \hat{V}_c(i) = \gamma^{i_0 - i},
        \end{equation}
        Then we can get the smoothness of both functions:
        \begin{equation}
            L(\hat{F}, \rho) = \E_{\rho} |\hat{F}(t) - \hat{F}(t+1)| = \frac{1}{T},
        \end{equation}
        \begin{equation}
            L(\hat{V}_c, \rho) = \E_{\rho} |\hat{V}_c(t) - \hat{V}_c(t+1)| = \frac{1}{T} (1 + \sum_{t=1}^{i_0-1}\gamma^{t-1}(1-\gamma)) = \frac{2-\gamma^{i_0-1}}{T} ,
        \end{equation}
        Then we can get that $L(\hat{V}_c, \rho) \geq L(\hat{F}, \rho)$.

        \item If exist two time steps $i_0 < i_1$ such that $c(i_0) = c(i_1) = 0$. 

        \begin{equation}
        \sum_{t=i_0}^{i_1-1} L(\hat{F}) = \hat{F}(i_0) - \gamma^{i_1-1} \hat{F}(i_1) + \sum_{t=i_0+1}^{i_1-1} \gamma^{t-1}(1-\gamma),
    \end{equation}
    Since $\hat{F}(i_0) = \hat{F}(i_1) = 1$, we can get:
    \begin{equation}
        \sum_{t=i_0}^{i_1-1} L(\hat{F}) = 2(1-\gamma^{i_1-i_0-1}).
    \end{equation}
    Then we turn to the smoothness score sum from time step $0$ to $t_0$ for the cost value function $\hat{V}_c$:
    \begin{equation}
    \begin{aligned}
        \sum_{t=i_0}^{i_1-1} L(\hat{V}_c) &= 1 + \gamma^{i_1 - i_0} \hat{V}_c(i_1) - \gamma^{i_1 - i_0 -1} \hat{V}_c(i_1) + \sum_{t=i_0+1}^{i_1-1} (1-\gamma)\hat{V}_c(i_1) \\
         &= \hat{V}_c(i_1)(1-\gamma^{t_0-1}) + 1 + \hat{V}_c(i_1)(\gamma^{i_1 - i_0} - \gamma^{i_1 - i_0-1})  \\ 
         &= 1 + \hat{V}_c(i_1)(1-2\gamma^{i_1 - i_0-1} + \gamma^{i_1 - i_0})
    \end{aligned}
    \end{equation}
    Then since $\hat{V}_c(t_0) \geq \hat{F}(t_0) = 0$, we can derive that:
    \begin{equation}
    \begin{aligned}
        \sum_{t=i_0}^{i_1-1} L(\hat{V}_c) &= 1 + \hat{V}_c(i_1)(1-2\gamma^{i_1 - i_0-1} + \gamma^{i_1 - i_0}) \\
        & \geq 1 + (1-2\gamma^{i_1 - i_0-1} + \gamma^{i_1 - i_0}) \\
        & = 2 (1-\gamma^{i_1 - i_0-1}) + \gamma^{i_1 - i_0}) \\
        & \geq 2 (1-\gamma^{i_1 - i_0-1}) =  \sum_{t=i_0}^{i_1-1} L(\hat{F}),
    \end{aligned}
    \end{equation}
    which means for the trajectory between $i_0$ and $i_1$, we have the averaged smoothness score order: $\E_{\rho[i_0:i_1]} \hat{F} \leq \E_{\rho[i_0:i_1]} \hat{V}_c$.

    \item For other conditions, the trajectories can be separated into sub-trajectories with conditions 1, 2, 3, then the smoothness order is $L(\hat{V}_c, \rho) \geq L(\hat{F}, \rho)$.
        
    \end{enumerate}

\end{proof}

\newpage
\section{Supplementary Materials for Experiments}
\subsection{Dynamics loss}
\label{app:dyn_loss}
For dynamics loss function $D^{\text{dyn}}(z_1,z_2)$ in eq.(\ref{eq:dynloss}), we use SimSiam~\cite{chen2021exploring} as self-supervision loss function. Specifically, we employ two trainable projection functions $p_1,p_2$ (modeled as 2-layer MLPs), the loss function is 
\begin{equation}
    \ell^{\text{dyn}}(z_1,z_2) = -\text{cosine}\langle p_2(p_1(z_1)), \text{sg}(p_1(z_2)) \rangle,
\end{equation}
where $\text{cosine}\langle\cdot,\cdot\rangle$ is the consine similarity and $\text{sg}$ means stopping gradient. During training, we will update $p_1,p_2$ along with the parameters of encoder, transition model, and prediction head $\theta$.

\subsection{Landscape}
\label{app:landscape}

\begin{figure}[htbp]
    \centering
    \begin{subfigure}[b]{0.33\columnwidth}
         \centering
         \includegraphics[width=0.8\textwidth]{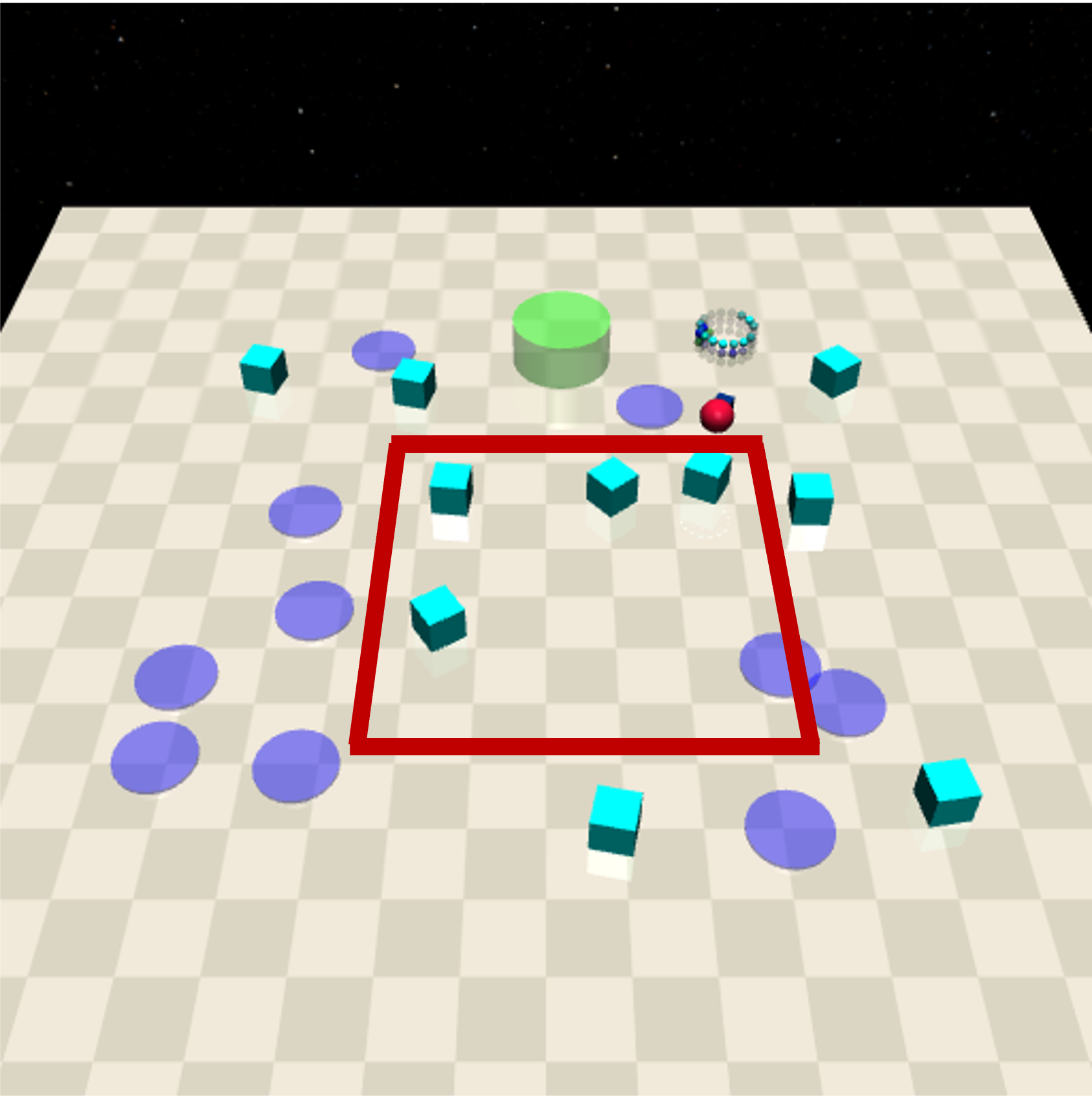}
         \caption{}
         % \label{fig:hm1}
     \end{subfigure}
     \hfill
    \begin{subfigure}[b]{0.33\columnwidth}
         \centering
         \includegraphics[width=1.0\textwidth]{figs/target_costvalue.png}
         \caption{}
         % \label{fig:hm1}
     \end{subfigure}
     \hfill
     \begin{subfigure}[b]{0.33\columnwidth}
         \centering
         \includegraphics[width=1.0\textwidth]{figs/target_feasibility.png}
         \caption{}
         % \label{fig:hm1}
     \end{subfigure}
     \caption{(a): The overview of the tested PointGoal2 task. Red point is the agent, blue squares are the obstacles, blue circles are hazards, and green cylinder is the goal, the agent should reach the goal with colliding with obstacles or stepping into hazards. We test the value and feasibility score of agent in the same positions in the \textit{red bounding box}. (b)(c): The landscape of target cost value and feasibility score obtained by bootstrapping in PointGoal2 task. The values of them have been rescaled.}
     \label{fig:landscape2}
\end{figure}

The landscape shown in fig.~\ref{fig:landscape2} corresponds to the red frame in the bird-view of the task. From the figure, we can observe that the landscape of value function is less smoother than feasibility score. Meanwhile, the peaks in landscape of feasibility score correspond to the four blue square obstacles, which shows a better match than value function. We also conduct a comparison of the embeddings learned by value consistent model and feasibility consistent model in Appendix~\ref{app:compare_vc_fc}.

\subsection{Comparison between the embedding learned by Value Consistent model and FCSRL}
\label{app:compare_vc_fc}

To further compare the learned embeddings by VC model and FCSRL, we test their quality by measuring the capability of predicting cost value $V_c(s)$ and feasibility score $F_c(s)$. Specifically, we 
\begin{enumerate}
    \item Train a SALE policy on PointGoal2 task (vector state) and sample 50 trajectories by SALE policy, store them to buffer $B$.
    \item Train a VC policy and FCSRL policy seperately, denote their encoder as $g_v$ and $g_f$; obtain corresponding embeddings of states in buffer $B$ as $\{z_v\}, \{z_f\}$; obtain target cost value $v_c^{(m)}$ and target feasibility $f_c^{(m)}$ by bootstrap estimate, respectively by VC and FCSRL. 
    \item Use linear regression models to train 4 models (input: $z_v$ or $z_f$; output $v_c$ or $f_c$); record the final MSE.
    \item Repeat above for several seeds.
\end{enumerate}
The final results are reported in table~\ref{tab:linear_mse}. We can find that FCSRL actually achieves similar MSE to VC model on cost value prediction although VC model has explicitly regressed the embedding to $V_c$ during training. Meanwhile, FCSRL has smaller MSE on feasibility score prediction. We also plot the prediction results on $V_c$ in fig.~\ref{fig:confusion_mat}. The comparison shows that the value function information is harder to extract via representation learning and the effectiveness of value consistent model is not very remarkable. In contrast, the feasibility score is easier to learned.

% \usepackage{booktabs}
\begin{table}[htb]
\centering
\begin{minipage}{0.56\linewidth}
\centering
\caption{The prediction MSE of embeddings from FCSRL and VC model.}
\label{tab:linear_mse}
\begin{tabular}{@{}lcc@{}}
\toprule
                       & MSE of cost value & MSE of feasibility \\ \midrule
FCSRL                  & 17.6±5.2                     & 0.139±0.047                   \\
VC model & 17.3±4.6                     & 0.217±0.096                   \\ \bottomrule
\end{tabular}
\end{minipage}\hfill
\begin{minipage}{0.4\linewidth}
\centering
\includegraphics[width=0.7\textwidth]{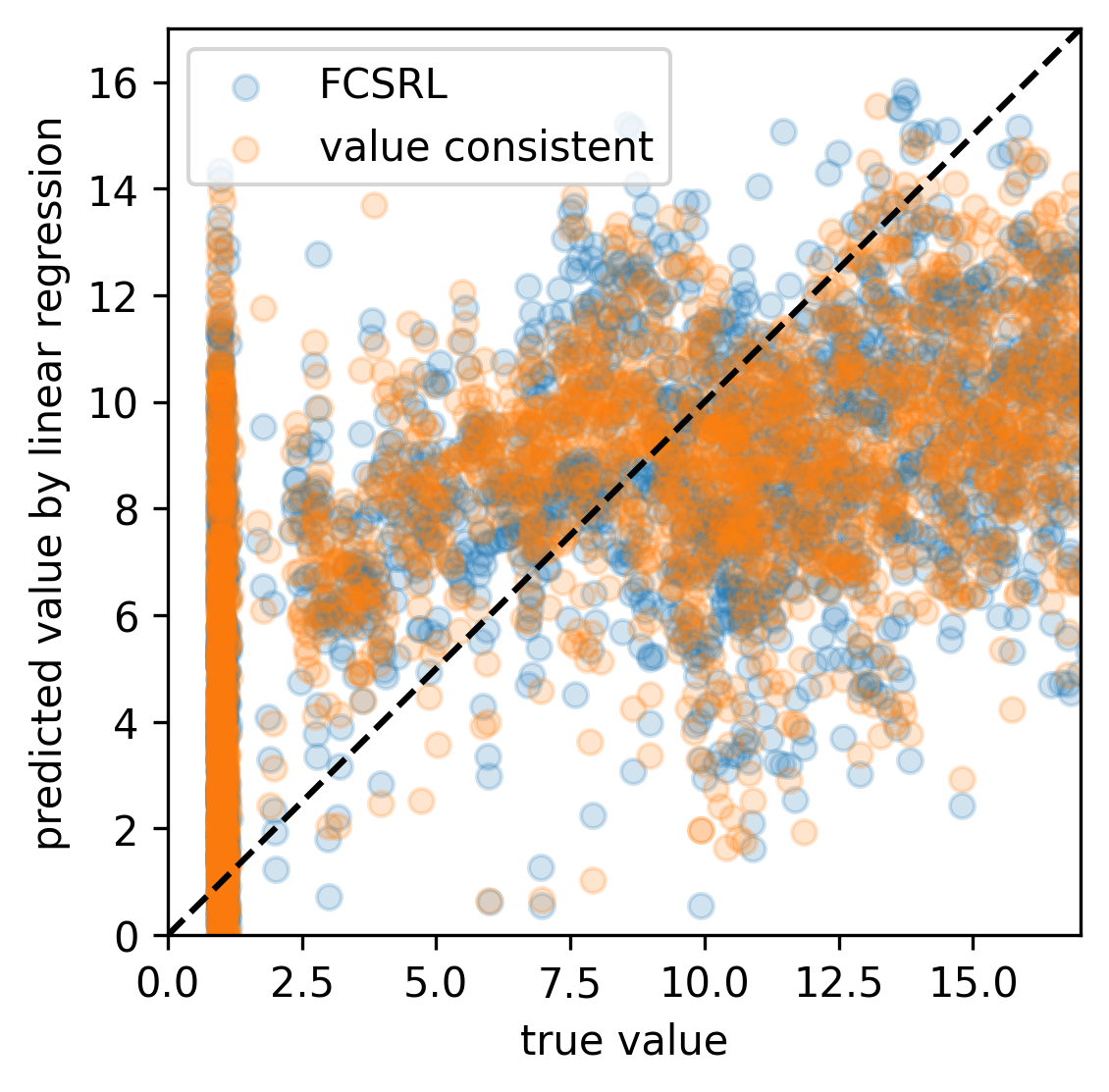}
\captionof{figure}{The comparison of VC model and FCSRL on prediction of $V_c$.}
\label{fig:confusion_mat}
\end{minipage}
\end{table}

% \subsection{More experiments}
% \label{app:more_exp}
% bullet safety gym
% more baselines

\subsection{Comparison with other safe RL baselines}
We add a comparison of FCSRL and other safe RL baselines on Safety-Gymnasium tasks. We adopt FSRL\footnote{\url{https://github.com/liuzuxin/FSRL/tree/main}} as the implementations of baselines to test their performances. Our method FCSRL adopts TD3-Lag as base algorithm. The cost threshold $\epsilon$ is 10 for all tasks.

\begin{table}[htb]
\centering
\caption{The value is averaged over 5 seeds. \textbf{Bold} value indicates the algorithm can roughly satisfy the constraint (cost return $\leq$ 11). \blue{Blue} value indicates the highest rewards ($\geq 0.95*$ highest reward) when satisfying the constraint.}
\label{tab:comparison_safety_gymnasium}
\resizebox{0.99\linewidth}{!}{
\begin{tabular}{@{}l|cc|cc|cc|cc|cc|cc@{}}
\toprule
\multicolumn{1}{c|}{\multirow{2}{*}{Method}} & \multicolumn{2}{c|}{PointGoal1} & \multicolumn{2}{c|}{PointButton1} & \multicolumn{2}{c|}{PointPush1} & \multicolumn{2}{c|}{PointGoal2} & \multicolumn{2}{c|}{CarGoal1} & \multicolumn{2}{c}{CarButton1} \\
\multicolumn{1}{c|}{}                        & reward         & cost           & reward          & cost            & reward         & cost           & reward        & cost            & reward        & cost          & reward        & Cost           \\ \midrule
CPO [\cite{achiam2017constrained}]                                          & \textbf{4.2}±1.3        & \textbf{10.8}±9.2       & -0.3±0.6        & 12.4±7.7        & 0.4±0.6        & 15.9±10.0      & -0.2±0.5      & 23.8±16.0       & 7.5±3.3       & 14.8±6.4      & 0.1±0.4       & \textbf{9.4}±3.5      \\
PPO-Lag [\cite{ray2019benchmarking}]                                     & \textbf{13.4}±1.6       & \textbf{10.2}±4.5       & \textbf{1.6}±1.4         & \textbf{9.5}±3.2       & \textbf{2.0}±0.9        & \textbf{9.6}±7.5       & 3.0±1.7       & 22.6±8.5       & 14.9±4.8     & 26.1±7.5     & -0.6±1.1       & 19.4±17.2      \\
% DDPG-Lag                                     & -1.1±0.7       & 33.6±28.7      & -4.0±2.6        & 24.5±13.6       & ±              & ±              & -0.5±0.5      & 43.8±32.6       & -0.9±2.1      & 39.2±23.0     & -2.0±1.7      & 24.8±22.7      \\
% TD3-Lag                                      & 13.5±5.8       & 66.6±26.4      & 5.9±4.6         & 35.8±22.5       & 0.6±0.5        & 23.9±17.2      & 4.0±5.5       & 111.3±39.0      & 12.5±3.1      & 38.1±18.0     & 1.8±1.6       & 43.7±19.9      \\
FOCOPS [\cite{zhang2020first}]                                       & 10.8±4.0       & 12.7±9.6      & 6.6±3.4         & 42.6±18.0       & \textbf{0.6}±0.3        & \textbf{10.2}±7.6       & 6.7±3.7       & 73.6±39.8       & 15.0±2.2      & 36.2±6.9      & 0.6±1.7       & 17.2±19.6      \\
CVPO [\cite{liu2022constrained}]                                         & \textbf{5.5}±3.5        & \textbf{9.9}±7.7       & \textbf{0.7}±0.5         & \textbf{10.8}±7.6       & \textbf{3.3}±1.5        & \textbf{11.0}±5.6       & 0.2±0.6       & 35.6±33.2       & \textbf{5.8}±2.0       & \textbf{10.2}±6.6      & \textbf{-0.8}±0.7      & \textbf{7.3}±3.2      \\
FCSRL (Ours)                            & \blue{\textbf{24.4}±1.4}       & \blue{\textbf{9.3}±5.8}        & \blue{\textbf{10.5}±2.3}        & \blue{\textbf{10.1}±6.3}        & \blue{\textbf{11.8}±4.5}       & \blue{\textbf{9.7}±8.1}        & \blue{\textbf{13.5}±2.6}      & \blue{\textbf{9.7}±7.8}         & \blue{\textbf{27.6}±2.0}      & \blue{\textbf{9.9}±6.4}       & \blue{\textbf{3.6}±1.3}       & \blue{\textbf{10.4}±7.3}        \\ \bottomrule
\end{tabular}
}
\end{table}

\subsection{More details of experiment settings}
\label{app:exp_setting}

\subsubsection{The performances of random policy and unconstrained PPO policy}

We report the reward performance of random policy $R_l$ and unconstrained PPO policy $R_h$ (which is used to normalize reward in fig.~\ref{fig:main_result}) in table~\ref{tab:random}.
\begin{table}[htb]
\centering
\caption{The performances of random and unconstrained PPO policies.}
\label{tab:random}
\begin{tabular}{@{}lcc@{}}
\toprule
             & random policy $R_l$ & unconstrained PPO $R_h$ \\ \midrule
PointGoal1   & -0.35               & 23.3                    \\
PointButton1 & -0.01               & 17.6                    \\
PointPush1   & -0.41               & 3.1                     \\
PointGoal2   & -0.20               & 22.2                    \\
CarGoal1     & -2.08               & 24.8                    \\
CarButton1   & -1.38               & 9.0                     \\ \bottomrule
\end{tabular}
\end{table}

\subsubsection{Network structure}
We adopt 2-layer MLPs for all transition models, prediction heads, projection networks in SimSiam loss, including both our method and baselines, both vector-state and image-based tasks. Besides, for encoder, we still adopt a 2-layer MLP for state-vector tasks while using a 4-layer CNN for image-based tasks. For fair comparison, we adopt the same neural network architecture for all baselines and our method. 

\subsubsection{Hyperparameters}
We adopt the same hyperparameters for all tasks in the same domain. For the vector-input and image-based tasks, we also keep most hyperparameters the same. The hyperparameters are summaries in table~\ref{tab:hp}. More details can be found in codes provided in \url{https://github.com/czp16/FCSRL}.

\begin{table}[htb]
\caption{The hyperparameters adopted in experiments.}
\label{tab:hp}
\centering
\begin{tabular}{@{}cc@{}}
\toprule
Hyperparameter                                         & Value                   \\ \midrule
hidden layers of actor                                 & {[}256, 256{]}          \\
hidden layers of critic                                & {[}256, 256{]}          \\
hidden layers of transition $h$                        & {[}256, 256{]}          \\
hidden layers of encoder $g$ for vector tasks          & {[}256, 256{]}          \\
NN optimizer                                           & Adam                    \\
NN learning rate                                       & 3e-4                    \\
Number of bins in discrete regression                  & 63                      \\
discount factor $\gamma$                               & 0.99                    \\
prediction length $K$                                  & 4                       \\
PID cofficient for Lagrangian $[K_p, K_i, K_d]$        & {[}0.02, 0.005, 0.01{]} \\
dimension of embedding $z$                             & 64/128 for vector/image \\
soft update coefficient for target network             & 0.05                    \\
coefficient of feasibility loss $\lambda^{\text{fea}}$ & 2.0                     \\ \bottomrule
\end{tabular}
\end{table}

\subsection{Training curves}
\label{app:trainig_curve}

We attach the training curves below.

% \begin{subfigure}
%      \centering
%      \includegraphics[width=\textwidth]{figs/main_train_curve_ppo.png}
%      \caption{Base method: PPO-Lagrangian.}
%      % \label{fig:hm1}
% \end{subfigure}
% \begin{subfigure}
%      \centering
%      \includegraphics[width=\textwidth]{figs/main_train_curve_td3.png}
%      \caption{Base method: TD3-Lagrangian.}
%      % \label{fig:hm1}
%  \end{subfigure}

 \begin{figure}[htb]
    \centering
    \begin{subfigure}[b]{0.95\textwidth}
     \centering
     \includegraphics[width=1.0\textwidth]{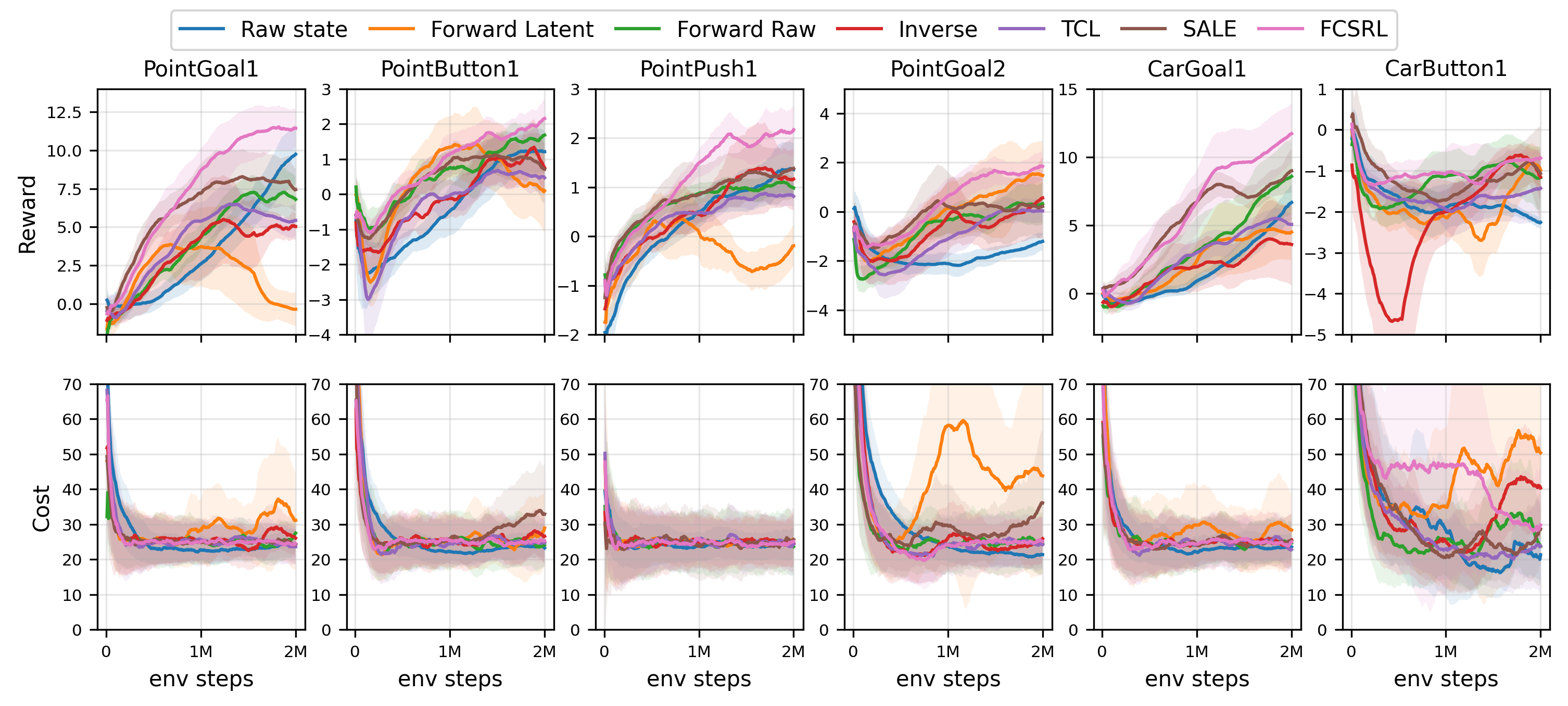}
     \caption{Base method: PPO-Lagrangian.}
     % \label{fig:hm1}
    \end{subfigure}
    
    \begin{subfigure}[b]{0.95\textwidth}
         \centering
         \includegraphics[width=1.0\textwidth]{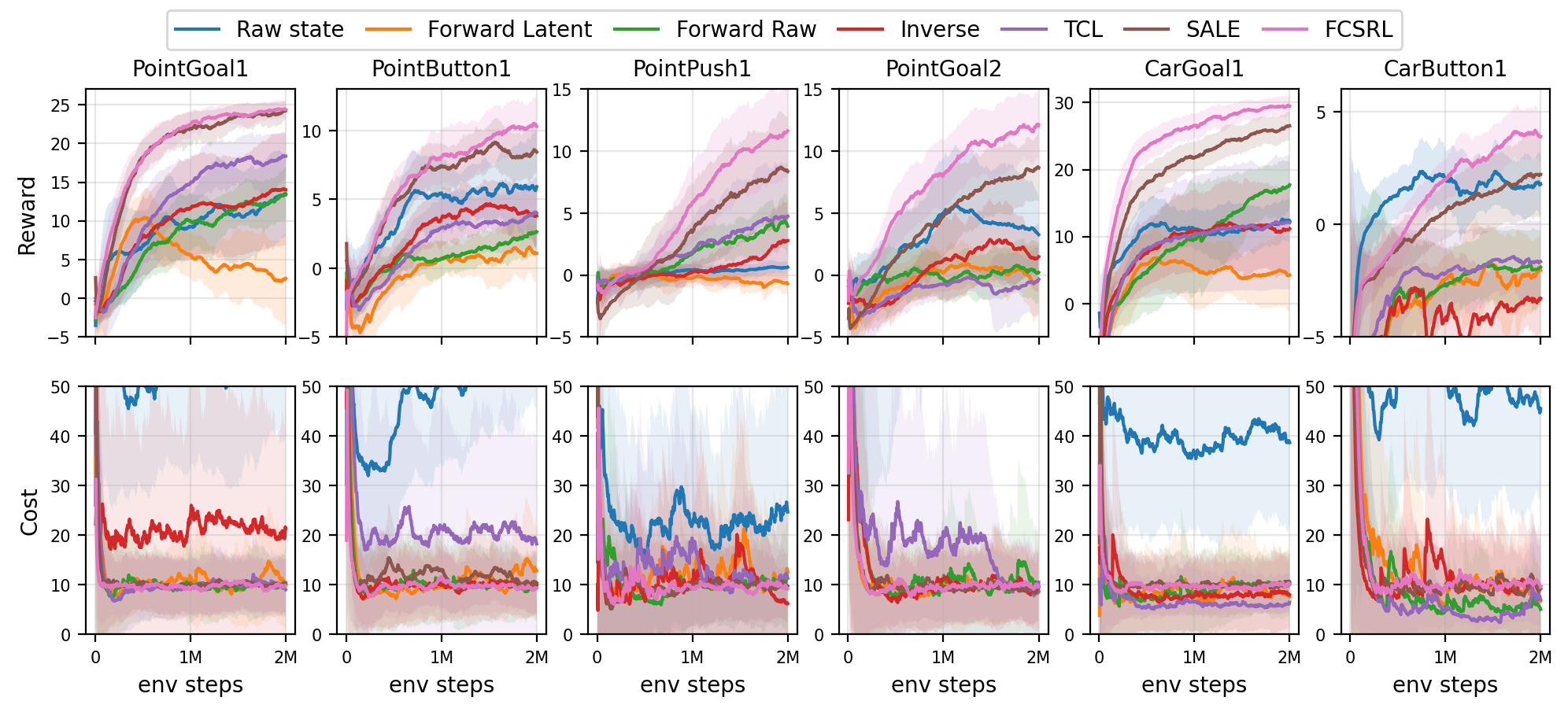}
         \caption{Base method: TD3-Lagrangian.}
         % \label{fig:hm1}
     \end{subfigure}
     \caption{The training curves of different representation learning methods based on PPO-Lagrangian and TD3-Lagrangian.}
\end{figure}

% \begin{figure}[htbp]
% \begin{floatrow}[2]
% \figurebox{\caption{ \small Optimization time}}{
% \centering
% \includegraphics[width=0.3\textwidth]{figs/main_train_curve_ppo.png}
% \label{fig: optimization time}
% }
% \figurebox{\caption{ \small Optimization time}}{
% \centering
% \includegraphics[width=0.3\textwidth]{figs/main_train_curve_td3.png}
% \label{fig: optimization time}
% }
% % \hspace{50pt}
% \end{floatrow}
    
% \end{figure}

% \subsection{Ablation of cost/cost value/feasibility}

\end{document}